\documentclass[10pt]{article} 
\usepackage[accepted]{tmlr}
\usepackage{rotating}
\usepackage{multirow}
\usepackage{algorithm}
\usepackage{subcaption}
\usepackage{graphicx}
\usepackage{amssymb}
\usepackage{amsthm}
\usepackage{hyperref}
\usepackage{booktabs}
\usepackage{url}
\usepackage{xcolor}
\usepackage{listings}
\usepackage{float}

\usepackage{amsmath,amsfonts,bm}

\def\eqref#1{equation~\ref{#1}}

\def\1{\bm{1}}

\DeclareMathAlphabet{\mathsfit}{\encodingdefault}{\sfdefault}{m}{sl}
\SetMathAlphabet{\mathsfit}{bold}{\encodingdefault}{\sfdefault}{bx}{n}

\DeclareMathOperator*{\argmax}{arg\,max}

\title{Uncertainty Quantification for Language Models: A Suite of Black-Box, White-Box, LLM Judge, and Ensemble Scorers}

\author{\name Dylan Bouchard \email dylan.bouchard@cvshealth.com \\
      \addr CVS Health, Wellesley, MA
      \AND
      \name Mohit Singh Chauhan \email mohitsingh.chauhan@cvshealth.com \\
      \addr CVS Health, Wellesley, MA}

\def\month{11}  
\def\year{2025} 
\def\openreview{\url{https://openreview.net/forum?id=WOFspd4lq5}} 

\begin{document}

\maketitle

\begin{abstract}
    Hallucinations are a persistent problem with Large Language Models (LLMs). As these models become increasingly used in high-stakes domains, such as healthcare and finance, the need for effective hallucination detection is crucial. To this end, we outline a versatile framework for closed-book hallucination detection that practitioners can apply to real-world use cases. To achieve this, we adapt a variety of existing uncertainty quantification (UQ) techniques, including black-box UQ, white-box UQ, and LLM-as-a-Judge, transforming them as necessary into standardized response-level confidence scores ranging from 0 to 1. To enhance flexibility, we propose a tunable ensemble approach that incorporates any combination of the individual confidence scores. This approach enables practitioners to optimize the ensemble for a specific use case for improved performance. To streamline implementation, the full suite of scorers is offered in this paper's companion Python toolkit, \texttt{uqlm}. To evaluate the performance of the various scorers, we conduct an extensive set of experiments using several LLM question-answering benchmarks. We find that our tunable ensemble typically surpasses its individual components and outperforms existing hallucination detection methods. Our results demonstrate the benefits of customized hallucination detection strategies for improving the accuracy and reliability of LLMs.

\end{abstract}

\section{Introduction}
\label{sec:intro}
Large language models (LLMs) are being increasingly used in production-level applications, often in high-stakes domains such as healthcare or finance. Consequently, there is a rising need to monitor these systems for the accuracy and factual correctness of model outputs. In these sensitive use cases, even minor errors can pose serious safety risks while also leading to high financial costs and reputational damage. A particularly concerning risk for LLMs is hallucination, where LLM outputs sound plausible but contain content that is factually incorrect. Many studies have investigated hallucination risk for LLMs (see \citet{huang2023surveyhallucinationlargelanguage, tonmoy2024comprehensivesurveyhallucinationmitigation, shorinwa2024surveyuncertaintyquantificationlarge, huang2024surveyuncertaintyestimationllms} for surveys of the literature). Even recent models, such as OpenAI's GPT-4.5, have been found to hallucinate as often as 37.1\% on certain benchmarks \citep{OpenAI}, underscoring the ongoing challenge of ensuring reliability in LLM outputs. 

Hallucination detection methods typically involve comparing ground truth texts to generated content, comparing source content to generated content, or quantifying uncertainty. Assessments that compare ground truth texts to generated content are typically conducted pre-deployment in order to quantify hallucination risk of an LLM for a particular use case. While important, this collection of techniques does not lend itself well to real-time evaluation and monitoring of systems already deployed to production. In contrast, techniques that compare source content to generated content or quantify uncertainty can compute response-level scores at generation time and hence can be used for real-time monitoring of production-level applications.

Uncertainty quantification (UQ) techniques can be used for hallucination detection in a closed-book setting, meaning they do not require access to a database of source content, ground truth texts, or internet access. These approaches are typically classified as either black-box UQ, white-box UQ, or LLM-as-a-Judge. Black-box UQ methods exploit the stochastic nature of LLMs and measure semantic consistency of multiple responses generated from the same prompt. White-box UQ methods leverage token probabilities associated with the LLM outputs to compute uncertainty or confidence scores. LLM-as-a-Judge methods use one or more LLMs to evaluate the factual correctness of a question-answer concatenation.

In this paper, we outline a versatile framework for generation-time, closed-book hallucination detection that practitioners can apply to real-world use cases.  To achieve this, we adapt a variety of existing black-box UQ, white-box UQ, and LLM-as-a-Judge methods, applying transformations as necessary to obtain standardized confidence scores that range from 0 to 1.\footnote{We employ UQ-style signals as a means to produce response-level confidence for hallucination detection, rather than pursuing uncertainty quantification as an end in itself. We do not separate aleatoric and epistemic uncertainty.} For improved customization, we propose a tunable ensemble approach that incorporates any combination of the individual scorers. The ensemble output is a simple weighted average of these individual components, where the weights can be tuned using a user-provided set of graded LLM responses. This approach enables practitioners to optimize the ensemble for a specific use case, leading to more accurate and reliable hallucination detection. Importantly, our ensemble is extensible, meaning practitioners can expand to include new components as research on hallucination detection evolves.

We evaluate the full suite of UQ scorers on responses generated by four LLMs across various question-answer benchmarks, yielding several empirical insights. Most notably, our tunable ensemble generally outperforms its individual components for hallucination detection. However, the performance ranking of individual scorers varies by dataset, underscoring the value of tailoring methods to specific use cases. Among black-box UQ scorers, entailment-style approaches demonstrate superior performance in our comparisons. Furthermore, gains from sampling additional responses shrink as the number of candidate responses rises, providing practical guidance for deployment. Lastly, a model’s accuracy on a given dataset positively relates to its performance as a judge of other models’ answers on that dataset.


Finally, this paper is complemented by our open-source Python package, \texttt{uqlm}, that provides ready-to-use implementations of all uncertainty quantification methods presented and evaluated in this work.\footnote{The \texttt{uqlm} repository can be found at \url{https://github.com/cvs-health/uqlm}. For a detailed description  of the software, we refer the reader to \cite{bouchard2025uqlmpythonpackageuncertainty}.} \texttt{uqlm}  enables practitioners to generate responses and obtain response-level confidence scores by providing prompts (i.e., the questions or tasks for the LLM) along with their chosen LLM. Our framework and toolkit provide researchers and developers a model-agnostic, user-friendly way to implement our suite of UQ-based scorers in real-world use cases, enabling more informed decisions around LLM outputs.

\section{Related Work}
\paragraph{Black-Box UQ}

\citet{cole2023selectivelyansweringambiguousquestions} propose evaluating similarity between an original response and candidate responses using exact match-based metrics. In particular, they propose two metrics: repetition, which measures the proportion of candidate responses that match the original response, and diversity, which penalizes a higher proportion of unique responses in the set of candidates. These metrics have the disadvantage of penalizing minor phrasing differences even if two responses have the same meaning. Text similarity metrics assess response consistency in a less stringent manner. \citet{manakul2023selfcheckgptzeroresourceblackboxhallucination} propose using n-gram-based evaluation to evaluate text similarity. Similar metrics such as ROUGE \citep{lin-2004-rouge}, BLEU \citep{10.3115/1073083.1073135}, and METEOR \citep{banerjee-lavie-2005-meteor} have also been proposed \citep{shorinwa2024surveyuncertaintyquantificationlarge}. These metrics, while widely adopted, have the disadvantage of being highly sensitive to token sequence orderings and often fail to detect semantic equivalence when two texts have different phrasing. Sentence embedding-based metrics such as cosine similarity \citep{9194665}, computed using a sentence transformer such as Sentence-Bert \citep{reimers2019sentencebertsentenceembeddingsusing}, have also been proposed \citep{shorinwa2024surveyuncertaintyquantificationlarge}. These metrics have the advantage of being able to detect semantic similarity in a pair of texts that are phrased differently. In a similar vein, \citet{manakul2023selfcheckgptzeroresourceblackboxhallucination} propose using BERTScore \citep{zhang2020bertscoreevaluatingtextgeneration}, based on the maximum cosine similarity of contextualized word embeddings between token pairs in two candidate texts. 

Natural Language Inference (NLI) models are another popular method for evaluating similarity between an original response and candidate responses. These models classify a pair of texts as either \textit{entailment}, \textit{contradiction}, or \textit{neutral}. Several studies propose using NLI estimates of $1-P(\text{contradiction})$  or $P(\text{entailment})$ between the original response and a set of candidate responses to quantify uncertainty \citep{chen2023quantifyinguncertaintyanswerslanguage, lin2024generatingconfidenceuncertaintyquantification}.  \citet{zhang2024luqlongtextuncertaintyquantification} follow a similar approach but instead average across sentences and exclude $P(\text{neutral})$ from their calculations.\footnote{Averaging across sentences is done to address long-form responses. \cite{jiang2024graphbaseduncertaintymetricslongform} also address long-form hallucination detection but follow a graph-based approach instead.} Other studies compute semantic entropy using NLI-based clustering \citep{kuhn2023semanticuncertaintylinguisticinvariances, kossen2024semanticentropyprobesrobust, Farquhar2024}. \citet{qiu2024semanticdensityuncertaintyquantification} estimate density in semantic space for candidate responses. 

\paragraph{White-Box UQ}

\cite{manakul2023selfcheckgptzeroresourceblackboxhallucination} consider two scores for quantifying uncertainty with token probabilities: average negative log probability and maximum negative log probability. While these approaches effectively represent a measure of uncertainty, they lack ease of interpretation, are unbounded, and are more useful for ranking than interpreting a standalone score. \cite{fadeeva2024factcheckingoutputlargelanguage} consider perplexity, calculated as the exponential of average negative log probability. Similar to average negative log probability, perplexity also has the disadvantage of being unbounded. They also consider response improbability, computed as the complement of the joint token probability of all tokens in the response. Although this metric is bounded and easy to interpret, it penalizes longer token sequences relative to semantically equivalent, shorter token sequences. Another popular metric is entropy, which considers token probabilities over all possible token choices in a pre-defined vocabulary \citep{malinin2021uncertaintyestimationautoregressivestructured, manakul2023selfcheckgptzeroresourceblackboxhallucination}. \cite{malinin2021uncertaintyestimationautoregressivestructured} also consider the geometric mean of token probabilities for a response, which has the advantage of being bounded and easy to interpret.\footnote{For additional white-box uncertainty quantification techniques, we refer the reader to \citet{ling2024uncertaintyquantificationincontextlearning, bakman2024marsmeaningawareresponsescoring, guerreiro2023lookingneedlehaystackcomprehensive, zhang2023enhancinguncertaintybasedhallucinationdetection, varshney2023stitchtimesavesnine, luo2023zeroresourcehallucinationpreventionlarge, ren2023selfevaluationimprovesselectivegeneration, vanderpoel2022mutualinformationalleviateshallucinations, wang2023selfconsistencyimproveschainthought}.}

\paragraph{LLM-as-a-Judge}

For uncertainty quantification, several studies concatenate a question-answer pair and ask an LLM to score or classify the answer's correctness. \citet{chen2023quantifyinguncertaintyanswerslanguage} propose using an LLM for self-reflection certainty, where the same LLM is used to judge correctness of the response. Specifically, the LLM is asked to score the response as incorrect, uncertain, or correct, which map to scores of 0, 0.5, and 1, respectively. Similarly, \citet{kadavath2022languagemodelsmostlyknow} ask the same LLM to state $P(Correct)$ given a question-answer concatenation. \citet{xiong2024llmsexpressuncertaintyempirical} explore several variations of similar prompting strategies for LLM self-evaluation. More complex variations such as multiple choice question answering generation \citep{manakul2023selfcheckgptzeroresourceblackboxhallucination}, multi-LLM interaction \citep{cohen2023lmvslmdetecting}, and follow-up questions \citep{agrawal2024languagemodelsknowtheyre} have also been proposed.

\paragraph{Ensemble Approaches}
\citet{chen2023quantifyinguncertaintyanswerslanguage} propose a two-component ensemble for zero-resource hallucination known as BSDetector. The first component, known as observed consistency, computes a weighted average of two comparison scores between an original response and a set of candidate responses, one based on exact match, and another based on NLI-estimated contradiction probabilities. The second component is self-reflection certainty, which uses the same LLM to judge correctness of the response. In their ensemble, response-level confidence scores are computed using a weighted average of observed consistency and self-reflection certainty. \cite{verga2024replacingjudgesjuriesevaluating} propose using a Panel of LLM evaluators (PoLL) to assess LLM responses. Rather than using a single large LLM as a judge, their approach leverages a panel of smaller LLMs. Their experiments find that PoLL outperforms large LLM judges, having less intra-model bias in the judgments.

\section{Hallucination Detection Methodology}

\subsection{Problem Statement}
We aim to model the binary classification problem of whether an LLM response contains a hallucination, which we define as any content that is nonfactual. To this end, we define a collection of binary classifiers, each of which map an LLM response $y_i \in \mathcal{Y}$, generated from prompt $x_i$, to a `confidence score' between 0 and 1, where $ \mathcal{Y}$ is the set of possible LLM outputs. We denote a hallucination classifier as $\hat{s}: \mathcal{Y} \xrightarrow[]{} [0, 1].$

Given a classification threshold $\tau$, we denote binary hallucination predictions from the classifier as $\hat{h}: \mathcal{Y} \xrightarrow{} \{0, 1 \}$. In particular, a hallucination is predicted if the confidence score is less than the threshold $\tau$:
\begin{equation}
    \hat{h}(y_i; \cdot, \tau) = \mathbb{I}(\hat{s}(y_i; \cdot) < \tau),
\end{equation}
where $\hat{s}$, and consequently $\hat{h}$, are conditioned on context variables including the prompt $x_i$ and, when applicable, additional responses generated from $x_i$.\footnote{For notational simplicity, we use ``$\cdot$'' as a placeholder here and explicitly write out conditioning variables in the scorer definitions that follow.} Note that $\hat{h}(\cdot)=1$ implies a hallucination is predicted. We denote the corresponding ground-truth label indicating whether the original response $y_i$ actually contains a hallucination as $h(y_i; \cdot)$. The function $h$ represents a grading process that compares a generated response $y_i$ to a correct reference answer $y_i^*$ for the same prompt $x_i$:

\begin{equation}
h(y_i; y_i^*, x_i) =   \begin{cases} 
          1 & y_i \text{ is incorrect with respect to } y_i^*\\
          0 & \text{otherwise}.
       \end{cases}
\end{equation}

When the correct answer $y_i^*$ is known, direct comparison yields the most accurate hallucination label. In deployment, however, $y_i^*$ is not available at generation time, so $h$ cannot be used as a real-time classifier. We therefore treat $h$ as an oracle ground truth used only offline for evaluation and, where applicable, tuning. Our framework approximates $h$ with $\hat{h}$ using uncertainty-based signals available at generation time, consistent with our closed-book setting and without access to $y_i^*$.


We adapt techniques from the uncertainty-quantification literature to compute response-level confidence for generation-time hallucination detection. Acknowledging the distinction between uncertainty quantification and hallucination detection, we study UQ-based confidence scorers for this purpose and do not distinguish between aleatoric and epistemic uncertainty. We transform and normalize outputs, if necessary, so that each score lies in $[0,1]$ with higher values indicating greater confidence.\footnote{Many scorers already output values in $[0,1]$ and therefore do not require normalization.} Below, we detail these scorers.

\subsection{Black-Box UQ Scorers}
\label{sec:using_multiple}
 Black-box UQ scorers exploit variation in LLM responses to the same prompt to assess semantic consistency. 
 For a given prompt $x_i$, these approaches involve generating $m$ candidate responses $\Tilde{\mathbf{y}}_i = \{ \Tilde{y}_{i1},...,\Tilde{y}_{im}\}$, using a non-zero temperature, from the same prompt and comparing these responses to the original response $y_{i}$. We provide detailed descriptions of each below.

\paragraph{Exact Match Rate.}
For LLM tasks that have a unique, closed-form answer, exact match rate can be a useful hallucination detection approach. Under this approach, an indicator function is used to score pairwise comparisons between the original response and the candidate responses. Given an original response $y_i$ and candidate responses $\Tilde{\mathbf{y}}_i$, generated from prompt $x_i$, exact match rate (EMR) is computed as follows:

\begin{equation}
    EMR(y_i; \Tilde{\mathbf{y}}_i, x_i) = \frac{1}{m} \sum_{j=1}^m \mathbb{I}(y_i=\Tilde{y}_{ij}).
\end{equation}

\paragraph{Non-Contradiction {Probability}.}
Non-contradiction probability (NCP) is a similar, but less stringent approach. NCP, a component of the BSDetector approach proposed by \citet{chen2023quantifyinguncertaintyanswerslanguage}, also conducts pairwise comparison between the original response and each candidate response. In particular, an NLI model is used to classify each pair $(y_{i}, \Tilde{y}_{ij})$ as \textit{entailment}, \textit{neutral}, or \textit{contradiction} and contradiction probabilities are saved. NCP for original response $y_{i}$ is computed as the average NLI-based non-contradiction probability across pairings with all candidate responses:
\begin{equation}
    NCP(y_i; \Tilde{\mathbf{y}}_i, x_i) = 1 - \frac{1}{m} \sum_{j=1}^m\frac{\eta(y_{i}, \Tilde{y}_{ij}) + \eta(\Tilde{y}_{ij},y_i)}{2}
\end{equation}

Above, $\eta(y_{i}, \Tilde{y}_{ij})$ denotes the contradiction probability of $(y_{i}, \Tilde{y}_{ij})$ estimated by the NLI model.\footnote{We note that NLI is an asymmetric measure. \citet{manakul2023selfcheckgptzeroresourceblackboxhallucination} propose a one-directional variant of NCP. We follow \citet{chen2023quantifyinguncertaintyanswerslanguage} in using the bidirectional formulation to allow for more flexibility in detecting contradictions.} Following \cite{chen2023quantifyinguncertaintyanswerslanguage} and  \cite{Farquhar2024}, we use \texttt{microsoft/deberta-large-mnli} for our NLI model.

\paragraph{BERTScore.}
Another approach for measuring text similarity between two texts is BERTScore \citep{zhang2020bertscoreevaluatingtextgeneration}. Let a tokenized text sequence be denoted as $\textbf{t} = \{t_1,...t_L\}$ and the corresponding contextualized word embeddings as $\textbf{E} = \{\textbf{e}_1,...,\textbf{e}_L\}$, where $L$ is the number of tokens in the text. The BERTScore precision and recall scores between two tokenized texts  $\textbf{t}, \textbf{t}'$ are respectively defined as follows:

\begin{equation}
    BertP(\textbf{t}, \textbf{t}') = \frac{1}{| \textbf{t}|} \sum_{t \in \textbf{t}} \max_{t' \in \textbf{t}'} \textbf{e} \cdot \textbf{e}' ; \quad BertR(\textbf{t}, \textbf{t}') = \frac{1}{| \textbf{t}'|} \sum_{t' \in \textbf{t}'} \max_{t \in \textbf{t}} \textbf{e} \cdot \textbf{e}'
\end{equation}

where $e, e'$ respectively correspond to $t, t'$. We compute our BERTScore confidence (BSC)  as follows:
\begin{equation}
    BSC(y_i; \Tilde{\mathbf{y}}_i, x_i) = \frac{1}{m} \sum_{j=1}^m 2\frac{ BertP(y_i, \Tilde{y}_{ij})  BertR(y_i, \Tilde{y}_{ij})}{BertP(y_i, \Tilde{y}_{ij})  + BertR(y_i, \Tilde{y}_{ij})},
\end{equation}
i.e. the average BERTScore F1 score across pairings of the original response with all candidate responses.

\paragraph{Normalized Cosine Similarity.}
Normalized cosine similarity (NCS) leverages a sentence transformer to map LLM outputs to an embedding space and measure similarity using those sentence embeddings. Let $V: \mathcal{Y} \xrightarrow{} \mathbb{R}^d$ denote the sentence transformer, where $d$ is the dimension of the embedding space. We define NCS as the average cosine similarity across pairings of the original response with all candidate responses, normalized by dividing by 2 and adding $\frac{1}{2}$:

\begin{equation}
    NCS(y_i; \Tilde{\mathbf{y}}_i, x_i) = \frac{1}{2m} \sum_{j=1}^m   \frac{\mathbf{V}(y_i) \cdot \mathbf{V}(\Tilde{y}_{ij}) }{ \lVert \mathbf{V}(y_i) \rVert \lVert \mathbf{V}(\Tilde{y}_{ij}) \rVert} + \frac{1}{2}.
\end{equation}

\paragraph{Normalized Semantic Negentropy.}
Semantic entropy (SE), proposed by \citet{Farquhar2024}, exploits variation in multiple responses to compute a measure of response volatility. The SE approach clusters responses by mutual entailment and, like the NCP scorer, relies on an NLI model. However, in contrast to the aforementioned black-box UQ scorers, semantic entropy does not distinguish between an original response and candidate responses. Instead, it computes a single metric value on a list of responses generated from the same prompt. We consider the discrete version of SE, defined  as follows:

\begin{equation}
    SE(y_i; \Tilde{\mathbf{y}}_i, x_i) = - \sum_{C \in \mathcal{C}} P(C|y_i, \Tilde{\mathbf{y}}_i)\log P(C|y_i, \Tilde{\mathbf{y}}_i),
\end{equation}
where $P(C|y_i, \Tilde{\mathbf{y}}_i)$ denotes the probability a randomly selected response $y \in \{ y_i, \Tilde{y}_{i1},...,\Tilde{y}_{im}\} $ belongs to cluster $C$, and $\mathcal{C}$ denotes the full set of clusters of $\{ y_i, \Tilde{y}_{i1},...,\Tilde{y}_{im}\}$.\footnote{If token probabilities of the LLM responses are available, the values of $P(C|y_i, \Tilde{\mathbf{y}}_i)$ can be instead estimated using mean token probability. However, unlike the discrete case, this version of semantic entropy is unbounded and hence does not lend itself well to normalization.} To ensure that we have a normalized confidence score with $[0,1]$ support and with higher values corresponding to higher confidence, we implement the following normalization to arrive at \textit{Normalized Semantic Negentropy} (NSN):
\begin{equation}
    NSN(y_i; \Tilde{\mathbf{y}}_i, x_i) = 1 - \frac{SE(y_i; \Tilde{\mathbf{y}}_i, x_i)}{\log (m + 1)},
\end{equation}
where $\log (m + 1)$ is included to normalize the support.

\subsection{White-Box UQ Scorers}
White-box UQ scorers leverage token probabilities of the LLM's generated response to quantify uncertainty.  We define two white-box UQ scorers below.

\paragraph{Length-Normalized Token Probability.}
Let the tokenization of LLM response $y_i$ be denoted as $\{t_1,...,t_{L_i}\}$, where $L_i$ denotes the number of tokens the response. Length-normalized token probability (LNTP) computes a length-normalized analog of joint token probability:

\begin{equation}
    LNTP(y_i; x_i) = \prod_{t \in y_i}  p_t^{\frac{1}{L_i}},
\end{equation}
where $p_t$ denotes the token probability for token $t$.\footnote{Although it is not reflected in our notation, the probability for a given token is conditional on the preceding tokens.} Note that this score is equivalent to the geometric mean of token probabilities for response $y_i$.

\paragraph{Minimum Token Probability.}
Minimum token probability (MTP) uses the minimum among token probabilities for a given responses as a confidence score:

\begin{equation}
    MTP(y_i; x_i) = \min_{t \in y_i}  p_t,
\end{equation}
where $t$ and $p_t$ follow the same definitions as above.

\subsection{LLM-as-a-Judge Scorers}
We employ an LLM-as-a-Judge scorer $J(y_i; x_i)$ as an additional method for obtaining response-level confidence scores. In this approach, we concatenate a question-response pair and pass it to an LLM with a carefully constructed instruction prompt that directs the model to evaluate the correctness of the response. We adapt our instruction prompt from \cite{xiong2024llmsexpressuncertaintyempirical}, instructing the LLM to score responses on a 0-100 scale, where a higher score indicates a greater certainty that the provided response is correct. These scores are then normalized to a 0-to-1 scale to maintain consistency with our other confidence scoring methods. The complete prompt template is provided in Appendix \ref{sec:judge_prompt}.

\subsection{Ensemble Scorer}
\label{sec:ensemble}
We introduce a tunable ensemble approach for hallucination detection. Specifically, our ensemble is a weighted average of $K$ binary classifiers: $\hat{s}_k: \mathcal{Y} \xrightarrow[]{} [0, 1]$ for $k=1,...,K$. As several of our ensemble components exploit variation in LLM responses to the same prompt, our ensemble is conditional on $(\Tilde{\mathbf{y}}_i, \mathbf{w})$, where $\mathbf{w}$ denote the ensemble weights. For original response $y_{i}$, we can write our ensemble classifier as follows:

\begin{equation}
    \hat{s}(y_i; \Tilde{\mathbf{y}}_i, x_i, \mathbf{w}) = \sum_{k=1}^K w_k \hat{s}_k(y_i; \Tilde{\mathbf{y}}_i, x_i),
\end{equation}
where $\mathbf{w}=(w_1,...,w_K), \sum_{k=1}^K w_k =1$, and $w_k \in [0,1]$ for $k=1,...,K$.\footnote{Note that although we write each classifier to be conditional on the set of candidate responses, some of the classifiers depend only on the original response.}

Tuning the ensemble requires a sample of LLM responses $y_1,...,y_n$ to a set of $n$ prompts, a set of $K$ confidence scores for each response $\{s_1(y_i; \Tilde{\mathbf{y}}_i, x_i),..., s_K(y_i; \Tilde{\mathbf{y}}_i, x_i)\}_{i=1}^N$, and corresponding binary hallucination indicators $h(y_1; y_1^*, x_1),...,h(y_n; y_n^*, x_n)$.\footnote{Note that $h(y_1; y_1^*, x_1),...,h(y_n; y_n^*, x_n)$ serve as `ground truth' labels in the classification objective function.} Given a classification objective function, the ensemble weights $\mathbf{w}$ can be tuned with an optimization routine.\footnote{We use \texttt{optuna} \citep{Akiba_Optuna_A_next-generation_2019} for optimization with default settings (more details available \href{https://optuna.readthedocs.io/en/stable/reference/generated/optuna.create_study.html}{here}).} If the objective is threshold-agnostic, the weights and threshold $\tau$ can be tuned sequentially. For a threshold-dependent objective, the weights and threshold can be tuned jointly. Because scorer performance depends on both the dataset and the underlying LLM, the ensemble weights are tuned per use case (chosen LLM and dataset), and the resulting weights are intended for in-domain deployment. See Appendix \ref{sec:tuning} for more details on ensemble tuning.

\section{Experiments}
\subsection{Experiment Setup} 
We conduct a series of experiments to assess the hallucination detection performance of the various scorers. To accomplish this, we leverage a set of publicly available benchmark datasets that contain questions and answers. To ensure that our answer format has sufficient variation, we use two benchmarks with numerical answers (\textit{GSM8K} \citep{cobbe2021trainingverifierssolvemath} and \textit{SVAMP} \citep{patel2021nlpmodelsreallyable}), two with multiple-choice answers (\textit{CSQA} \citep{talmor2022commonsenseqa20exposinglimits} and \textit{AI2-ARC} \citep{clark2018thinksolvedquestionanswering}), and two with open-ended text answers (\textit{PopQA} \citep{mallen-etal-2023-trust} and \textit{NQ-Open} \citep{lee2019latentretrievalweaklysupervised}). 

We sample 1000 questions for each of the six benchmarks. For each question, we generate an original response and $m=15$ candidate responses using four LLMs: GPT-4o \citep{OpenAI_doc}, GPT-4o-mini \citep{OpenAI_doc}, Gemini-2.5-Flash \citep{gemini_doc}, and Gemini-2.5-Flash-Lite
\citep{gemini_doc}.\footnote{We use a large number of candidate responses ($m=15$) to ensure robust comparisons across black-box scorers. In practice, fewer candidates can be used. For an experimental evaluation of the impact of $m$ on performance, refer to Appendix \ref{sec:num_responses}.} Candidate responses are generated using a temperature of 1.0. For each response, we use the corresponding candidate responses generated by the same LLM to compute the full suite of black-box UQ scores. We extract the log-probabilities from the generated responses to compute white-box UQ scores. Lastly, we compute LLM-as-a-Judge scores for each response using the same four LLMs. We evaluate the hallucination detection performance of all individual scorers as well as our ensemble scorer for each of the benchmarks using various metrics, with grading processes detailed in Appendix \ref{sec:grading}. All scores are computed using this paper's companion toolkit, \texttt{uqlm}.\footnote{Using an \texttt{n1-standard-16} machine (16 vCPU, 8 core, 60 GB memory) with a single NVIDA T4 GPU attached, our experiments took approximately 0.5-3 hours per LLM-dataset combination to complete.}

\subsection{Experiment Results}
\paragraph{Threshold-Agnostic Evaluation}
To assess the performance of the scorers as hallucination classifiers, we evaluate the performance of the confidence scores in a threshold-agnostic fashion.\footnote{In our experiments, we label `correct' LLM responses as 1 and `incorrect' responses as 0.} Under this setting, we use the AUROC-optimized ensemble weights and compute the ensemble's AUROC using 5-fold cross-validation.\footnote{Scorer performance depends on both the dataset and the LLM, so we tune the ensemble for a specific LLM–dataset pair. Our experiments reflect this in-domain setup; we do not evaluate out-of-distribution generalization across datasets or domains.}
 The final reported AUROC is obtained by averaging the AUROC values across the five holdout sets.

\begin{figure}[H]
    \begin{subfigure}{\textwidth}
    \centering
        \includegraphics[width=0.6\textwidth]{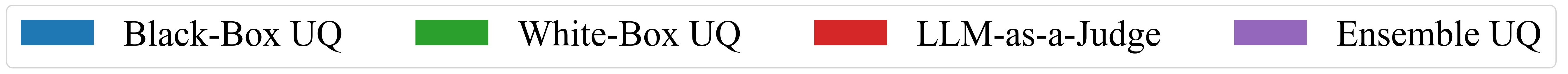}
    \end{subfigure}

    \centering
    \begin{subfigure}[b]{0.48\textwidth}
        \centering
        \includegraphics[width=\textwidth]{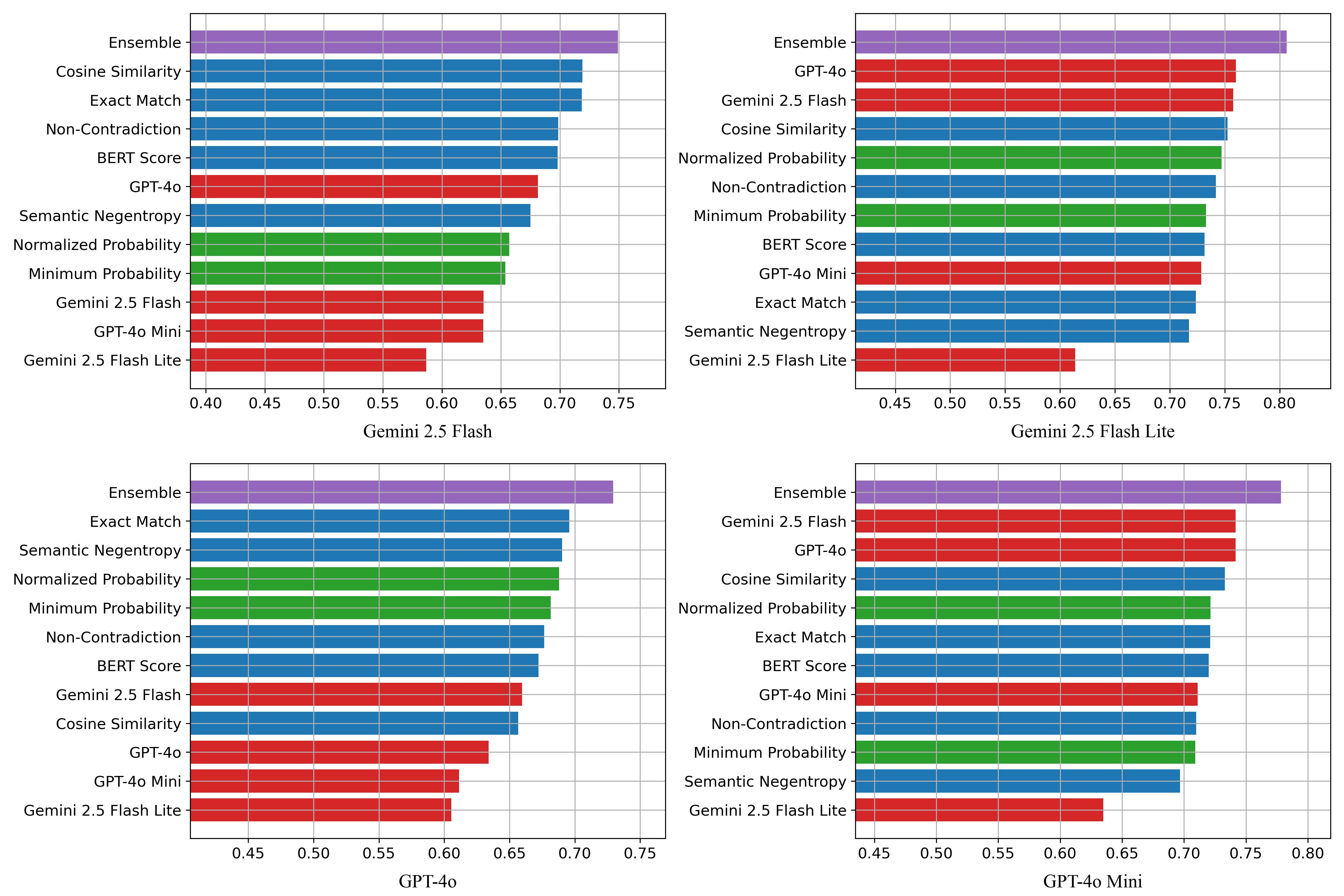}
        \caption{NQ Open}
        \label{fig:a}
    \end{subfigure}
    \hfill
    \begin{subfigure}[b]{0.48\textwidth}
        \centering
        \includegraphics[width=\textwidth]{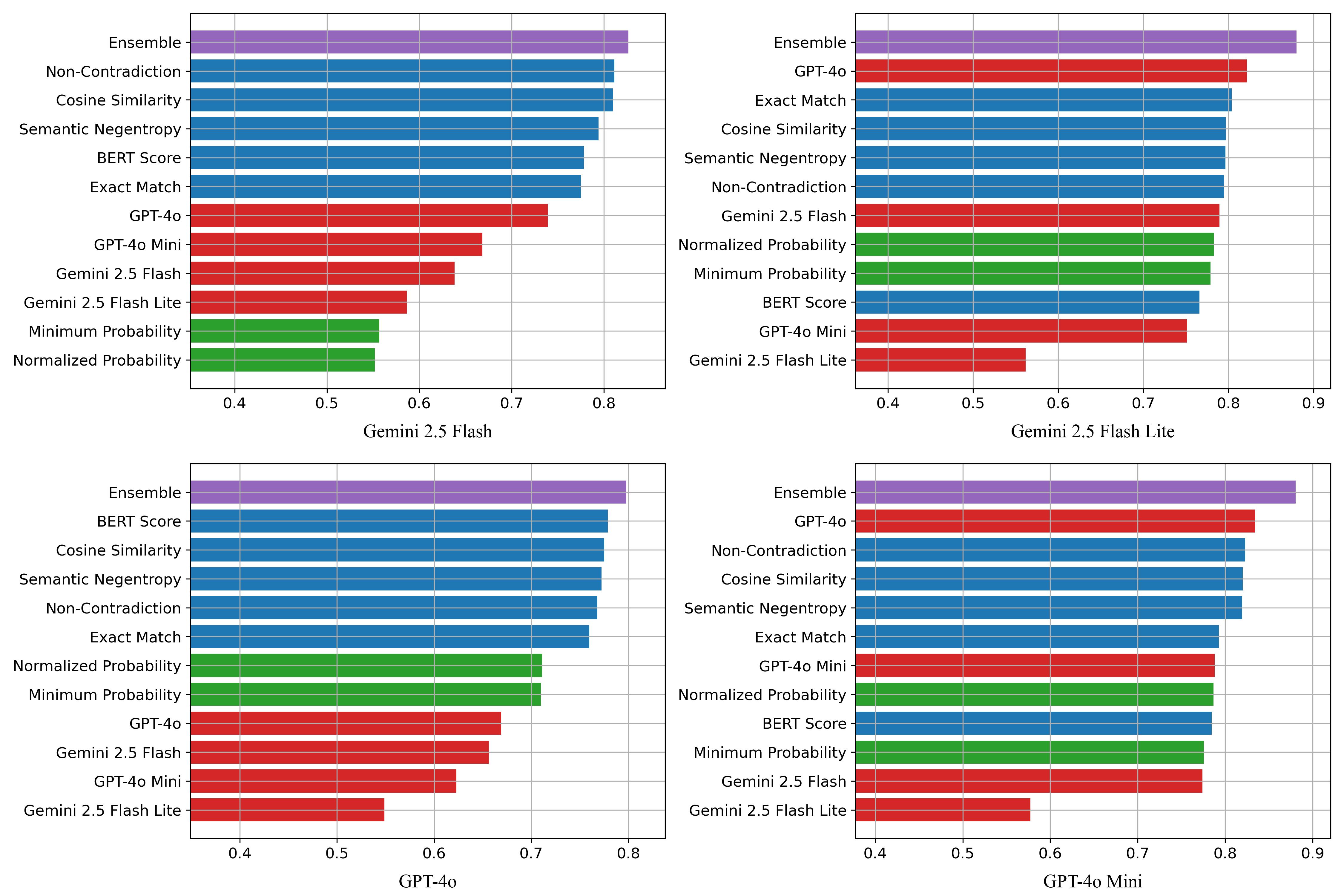}
        \caption{PopQA}
        \label{fig:b}
    \end{subfigure}

        \vspace{0.1cm}
    
    \begin{subfigure}[b]{0.48\textwidth}
        \centering
        \includegraphics[width=\textwidth]{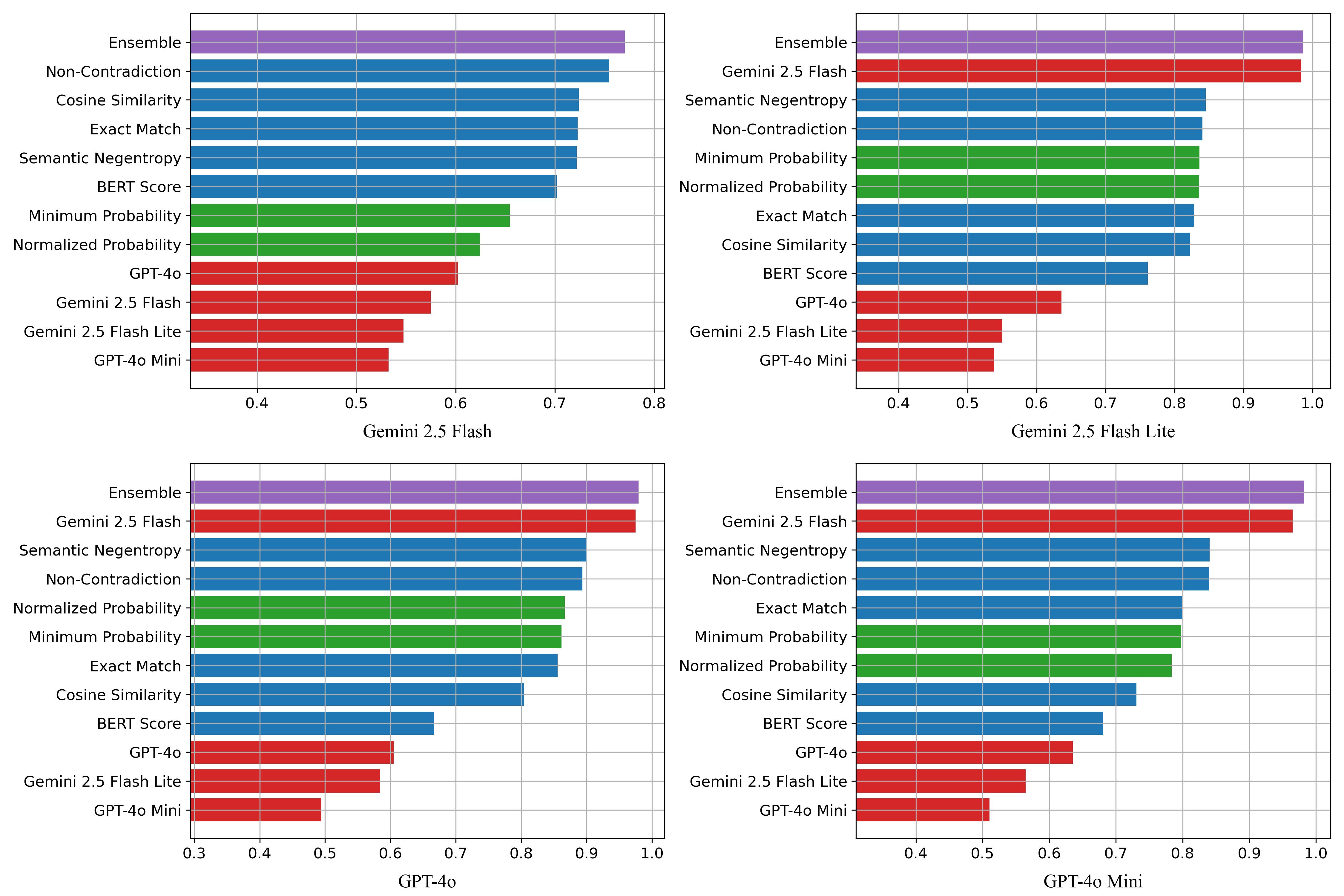}
        \caption{GSM8K}
        \label{fig:c}
    \end{subfigure}
    \hfill
    \begin{subfigure}[b]{0.48\textwidth}
        \centering
        \includegraphics[width=\textwidth]{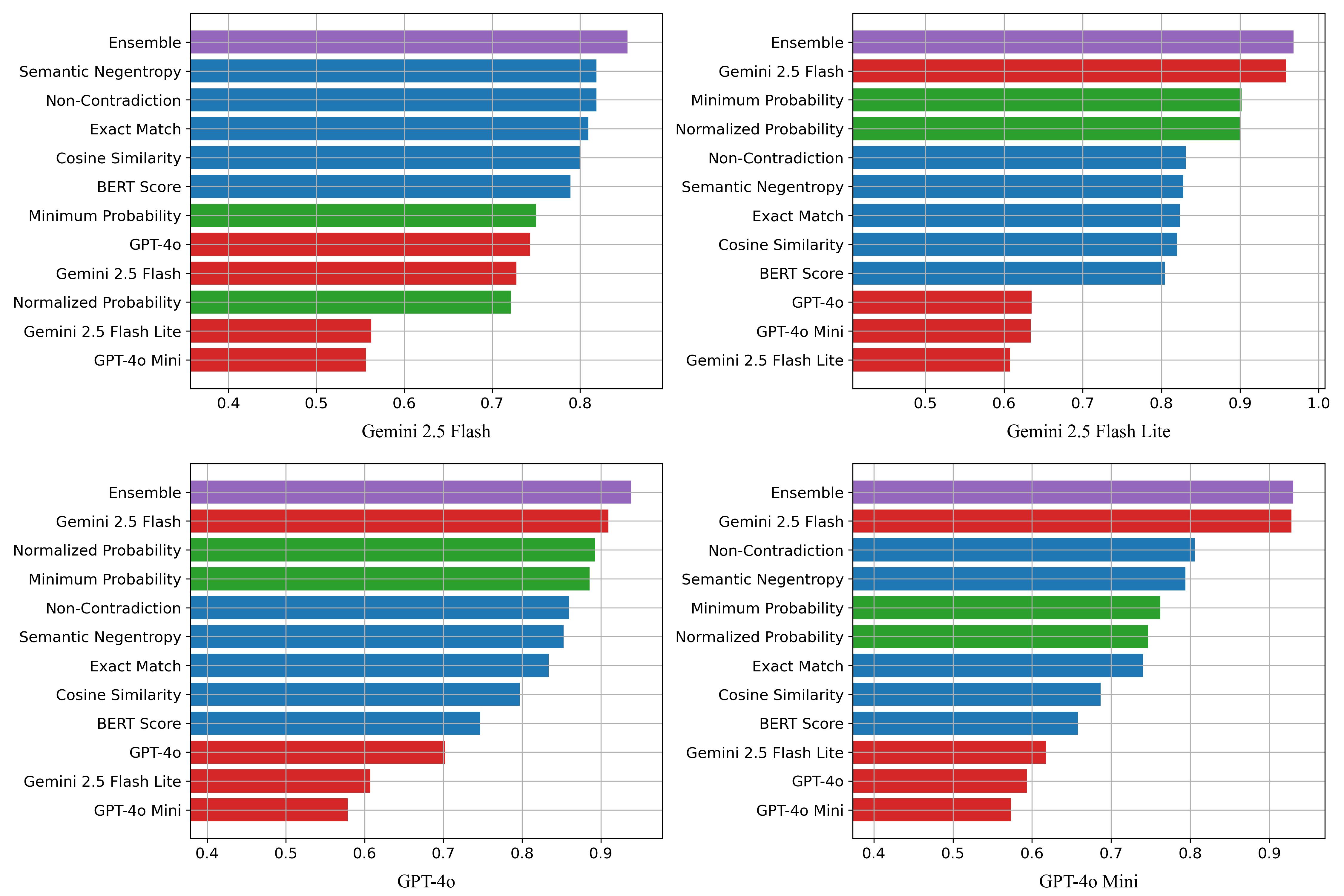}
        \caption{SVAMP}
        \label{fig:d}
    \end{subfigure}
    
    \vspace{0.1cm}
    
    \begin{subfigure}[b]{0.48\textwidth}
        \centering
        \includegraphics[width=\textwidth]{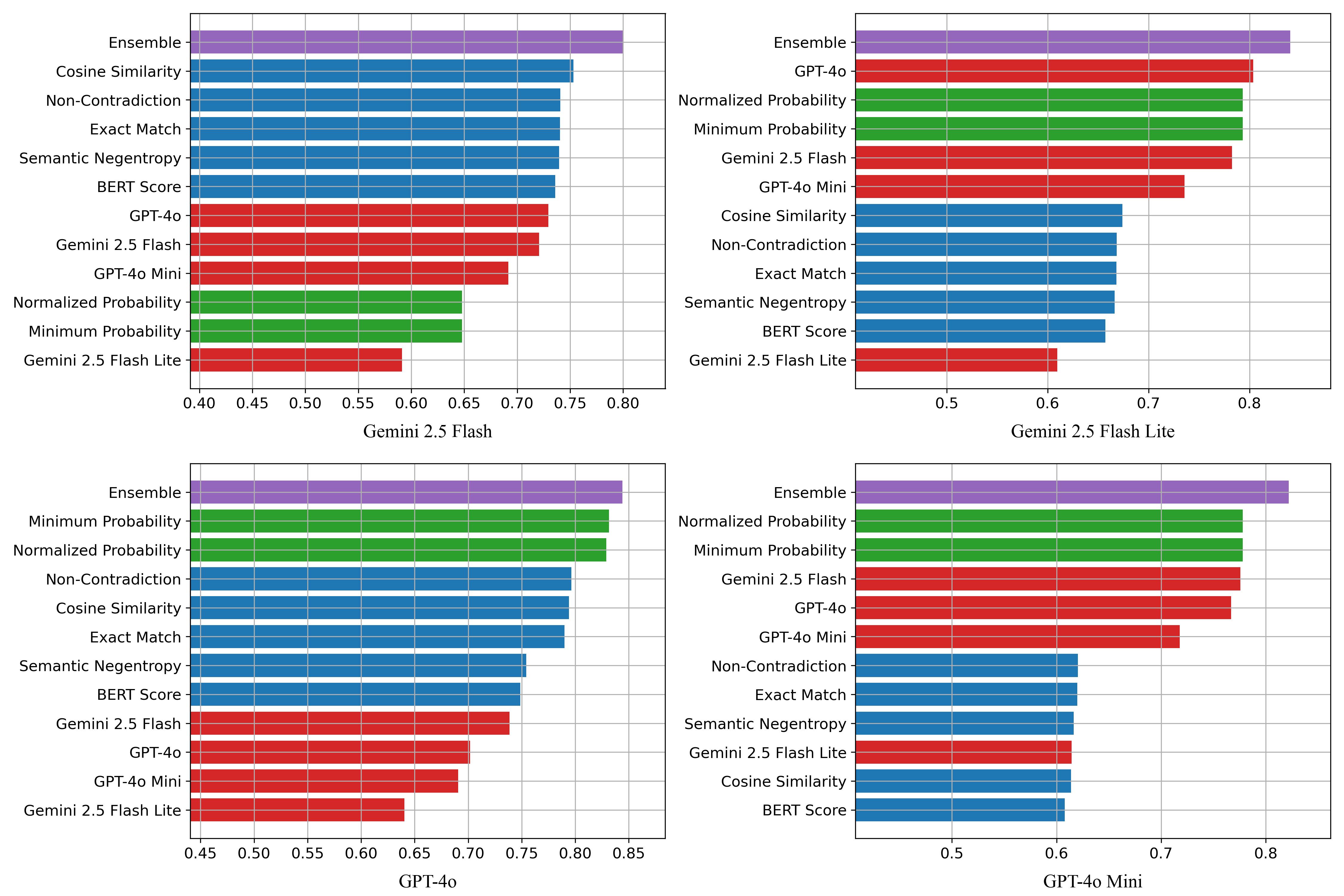}
        \caption{CSQA}
        \label{fig:e}
    \end{subfigure}
    \hfill
    \begin{subfigure}[b]{0.48\textwidth}
        \centering
        \includegraphics[width=\textwidth]{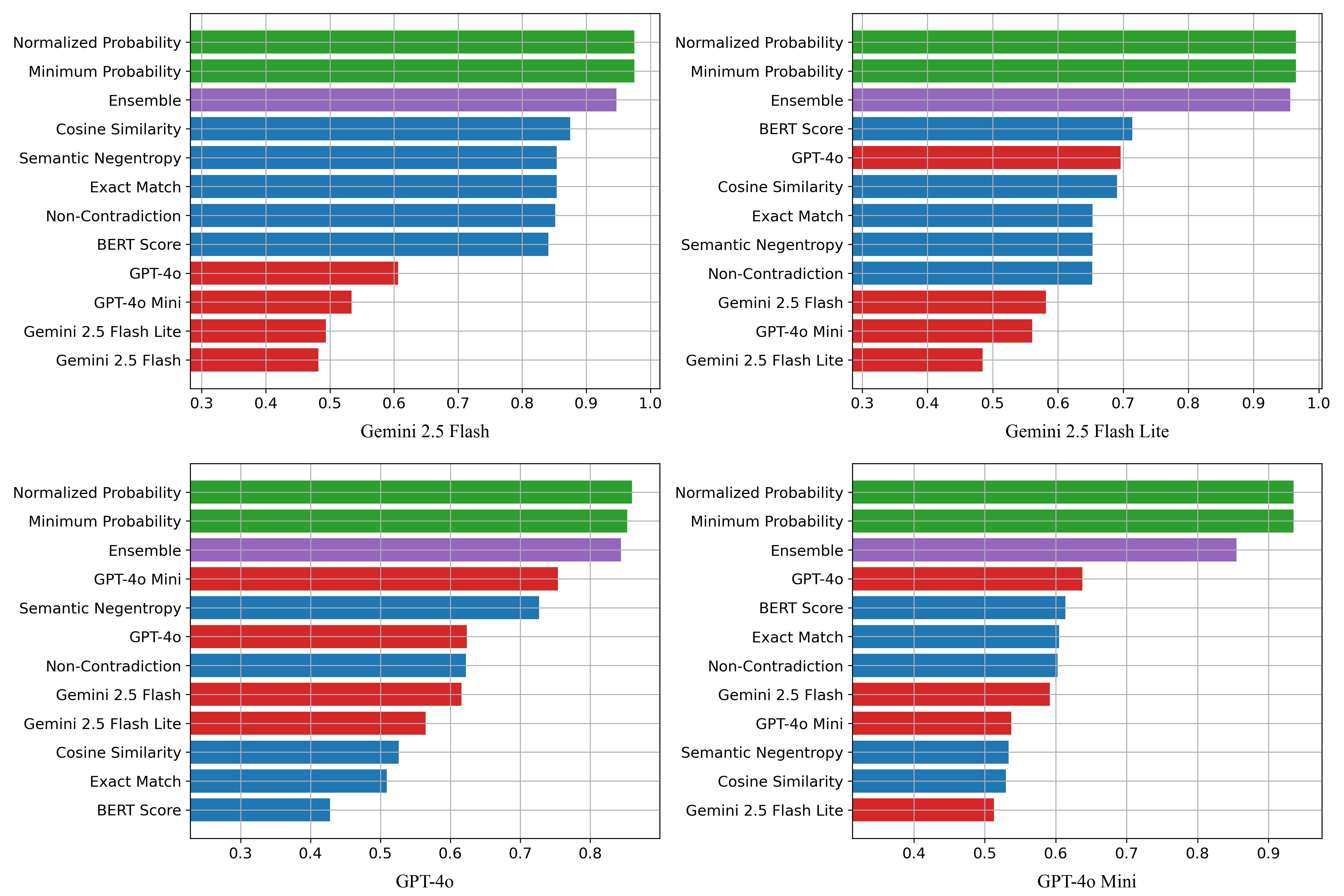}
        \caption{AI2-ARC}
        \label{fig:f}
    \end{subfigure}

    \caption{Scorer-Specific AUROC Scores for Hallucination Detection by LLM and Dataset (Higher is Better)}
    \label{fig:auroc}
\end{figure}

Figure \ref{fig:auroc} presents the AUROC scores for all scorers across the 24 LLM-dataset scenarios, while Table \ref{table:AUROC} highlights the best-performing scorer for each scenario. The AUROC values for the scenario-specific best scorers range from 0.729 for GPT-4o responses on NQ-Open (Ensemble) to 0.986 for Gemini-2.5-Flash-Lite responses on GSM8K (Ensemble). Overall, the top-performing scorers for each scenario exhibit strong hallucination detection performance, with AUROC values greater than 0.8 for 19 out of 24 scenarios. However, some scorers perform no better than or only negligibly better than a random classifier in hallucination detection in certain scenarios, such as the GPT-4o-mini LLM judge for all four GSM8K scenarios.

When comparing AUROC values across scorers, we find our ensemble scorer outperforms its individual components in 20 out of 24 scenarios, demonstrating the benefits of use-case-specific optimization. The rankings of individual scorers, however, vary significantly across scenarios, with LLM-as-a-Judge, black-box, and white-box methods achieving top non-ensemble performance in 11, 7, and 6 scenarios respectively. NLI-based scorers (NSN and NCP) exhibit strong performance, achieving the highest AUROC in 13 out of 24 scenarios among black-box scorers. Among LLM judges, GPT-4o and Gemini-2.5-Flash consistently outperform their smaller counterparts (GPT-4o-mini and Gemini-2.5-Flash-Lite), with GPT-4o ranking as the highest-performing judge in 13 out of 24 scenarios and Gemini-2.5-Flash in 10 out of 24 scenarios. Notably, Gemini-2.5-Flash tended to be the highest performing LLM judge on the math benchmarks (6 out of 8 scenarios), and GPT-4o tended to be the highest performing judge on the short-answer benchmarks (6 out of 8 scenarios). Finally, the two white-box scorers perform approximately equally, with very similar AUROC scores in the vast majority of scenarios.

\begin{table}[H]
\centering
\caption{Hallucination Detection AUROC (Higher is Better): Best-Performing Scorer by LLM and Dataset}
\small
\label{table:AUROC}
\begin{tabular}{cccccccc}
Model & Metric & NQ-Open & PopQA & GSM8K & SVAMP & CSQA & AI2-ARC \\
\midrule
\multirow{2}{*}{\begin{tabular}[c]{@{}c@{}}Gem.-2.5-Flash\end{tabular}}  & AUROC  & 0.749  & 0.826   & 0.771  & 0.854  & 0.800   & 0.975     \\
   & Best Scorer  & Ensemble  & Ensemble  & Ensemble   & Ensemble    & Ensemble  & LNTP \\
\midrule
\multirow{2}{*}{\begin{tabular}[c]{@{}c@{}}Gem.-2.5-Flash-Lite\end{tabular}} & AUROC  & 0.806  & 0.880   & 0.986  & 0.968  & 0.840   & 0.965     \\
   & Best Scorer  & Ensemble  & Ensemble  & Ensemble   & Ensemble   & Ensemble  & LNTP \\
\midrule
\multirow{2}{*}{\begin{tabular}[c]{@{}c@{}}GPT-4o\end{tabular}} & AUROC  & 0.729  & 0.798   & 0.979  & 0.938  & 0.844   & 0.860     \\
   & Best Scorer  & Ensemble  & Ensemble  & Ensemble   & Ensemble    & Ensemble  & LNTP \\
\midrule
\multirow{2}{*}{\begin{tabular}[c]{@{}c@{}}GPT-4o-Mini\end{tabular}} & AUROC  & 
0.778  & 0.880   & 0.982  & 0.930  & 0.822   & 0.935     \\
   & Best Scorer  & Ensemble  & Ensemble  & Ensemble   & Ensemble   & Ensemble  & LNTP \\
\bottomrule
\end{tabular}
\end{table}


\paragraph{Threshold-Optimized Evaluation.}

We evaluate the various scorers using a threshold-dependent metric (F1-score). To compute our ensemble scores in this setting, we jointly optimize the ensemble weights and threshold using F1-score as the objective function, as outlined in Appendix \ref{sec:tuning}. To ensure robust evaluations, we compute the scorer-specific F1-scores using 5-fold cross-validation. For each individual scorer, we select the F1-optimal threshold using grid search on the tuning set and compute F1-score on the holdout set. We report the final F1-score for each scorer as the average across holdout sets.

\begin{table}[H]
\centering
\caption{Hallucination Detection F1-Scores (Higher is Better): Best-Performing Scorer by LLM and Dataset}
\small
\label{table:F1}
\begin{tabular}{cccccccc}

Model & Metric & NQ-Open & PopQA & GSM8K & SVAMP & CSQA & AI2-ARC \\
\midrule
\multirow{4}{*}{\begin{tabular}[c]{@{}c@{}}Gem.-2.5-Flash\end{tabular}}  
& Precision  & 0.597  & 0.688   & 0.950  & 0.968  & 0.856   & 0.985     \\
& Recall  & 0.907  & 0.890   & 0.999  & 0.999  & 0.986   & 0.987     \\
& F1-Score & 0.718  & 0.774   & 0.974  & 0.983  & 0.916   & 0.986     \\
& Best Scorer & Ensemble  & Ensemble  & Ensemble   & Ensemble    & Ensemble  & NSN \\
\midrule
\multirow{4}{*}{\begin{tabular}[c]{@{}c@{}}Gem.-2.5-Fl.-Lt.\end{tabular}} 
& Precision  & 0.587  & 0.680   & 0.960  & 0.977  & 0.863   & 0.968     \\
& Recall  & 0.838  & 0.864   & 0.980  & 0.976  & 0.973   & 0.997     \\
& F1-Score & 0.688  & 0.759   & 0.969  & 0.977  & 0.915   & 0.982     \\
& Best Scorer & Ensemble  & Ensemble  & Ensemble   & Gem.-2.5-Fl.   & Ensemble  & Gem.-2.5-Fl.  \\
\midrule
\multirow{4}{*}{\begin{tabular}[c]{@{}c@{}}GPT-4o\end{tabular}} 
& Precision  & 0.649  & 0.727   & 0.958  & 0.978  & 0.853   & 0.987     \\
& Recall  & 0.917  & 0.880   & 0.987  & 0.990  & 0.972   & 1.000     \\
& F1-Score & 0.756  & 0.795   & 0.972  & 0.984  & 0.909   & 0.993     \\
& Best Scorer & Ensemble  & Ensemble  & Ensemble   & Gem.-2.5-Fl.    & NCP  & Gem.-2.5-Fl.  \\
\midrule
\multirow{4}{*}{\begin{tabular}[c]{@{}c@{}}GPT-4o-Mini\end{tabular}} 
& Precision  & 0.616  & 0.700   & 0.906  & 0.969  & 0.840   & 0.952     \\
& Recall  & 0.867  & 0.876   & 0.978  & 0.988  & 0.991   & 0.998     \\
& F1-Score & 0.718  & 0.776   & 0.940  & 0.978  & 0.909   & 0.975     \\
& Best Scorer & Ensemble  & Ensemble  & Ensemble   & Ensemble    & Ensemble  & Gem.-2.5-Fl.  \\
\bottomrule
\end{tabular}
\end{table}

\begin{figure}[]
    \begin{subfigure}{\textwidth}
    \centering
        \includegraphics[width=0.6\textwidth]{new_plots/legend.jpg}
    \end{subfigure}

    \centering
    \begin{subfigure}[b]{0.48\textwidth}
        \centering
        \includegraphics[width=\textwidth]{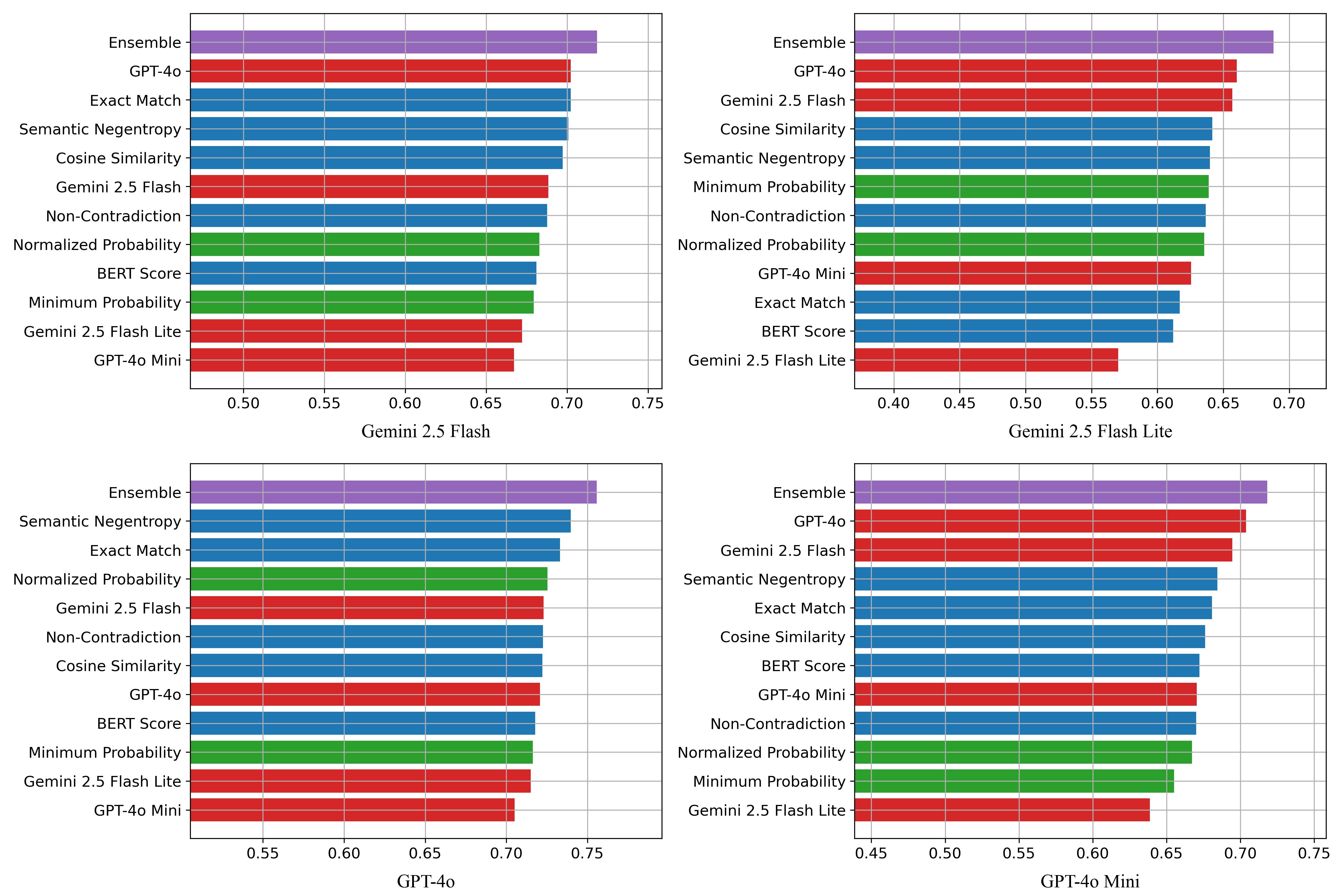}
        \caption{NQ Open}
        \label{fig:a}
    \end{subfigure}
    \hfill
    \begin{subfigure}[b]{0.48\textwidth}
        \centering
        \includegraphics[width=\textwidth]{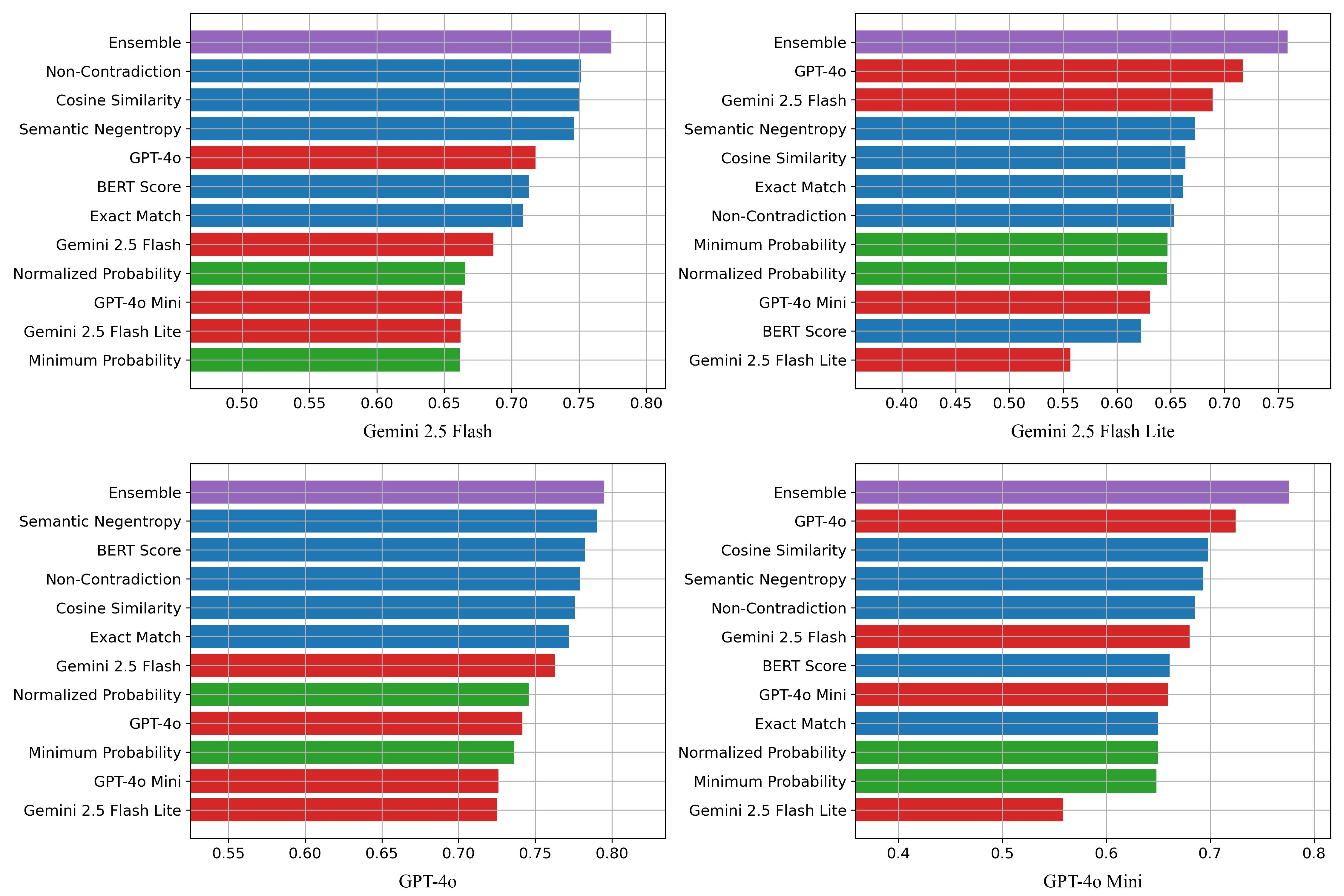}
        \caption{PopQA}
        \label{fig:b}
    \end{subfigure}

        \vspace{0.1cm}
    
    \begin{subfigure}[b]{0.48\textwidth}
        \centering
        \includegraphics[width=\textwidth]{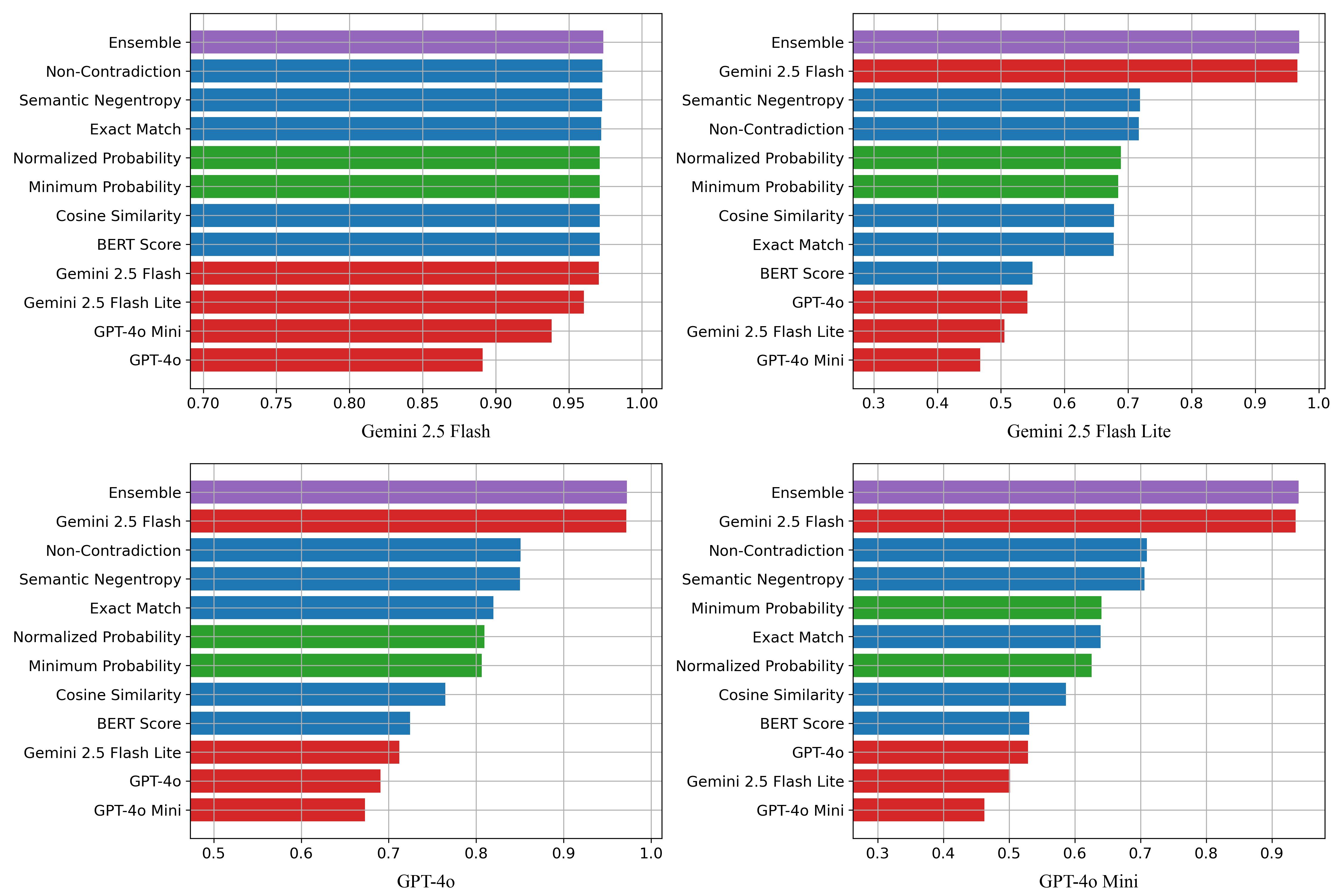}
        \caption{GSM8K}
        \label{fig:c}
    \end{subfigure}
    \hfill
    \begin{subfigure}[b]{0.48\textwidth}
        \centering
        \includegraphics[width=\textwidth]{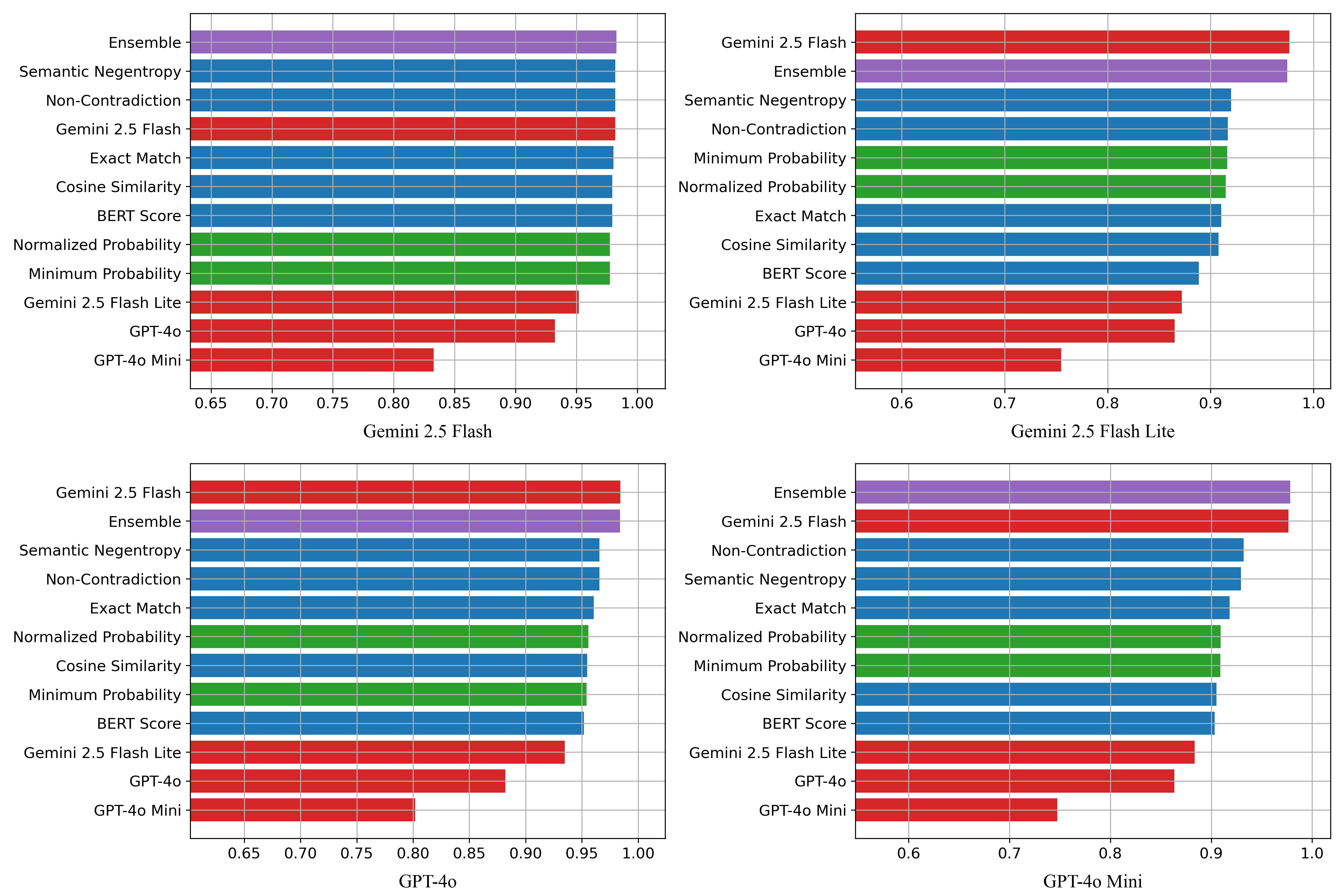}
        \caption{SVAMP}
        \label{fig:d}
    \end{subfigure}
    
        \vspace{0.1cm}
    
    \begin{subfigure}[b]{0.48\textwidth}
        \centering
        \includegraphics[width=\textwidth]{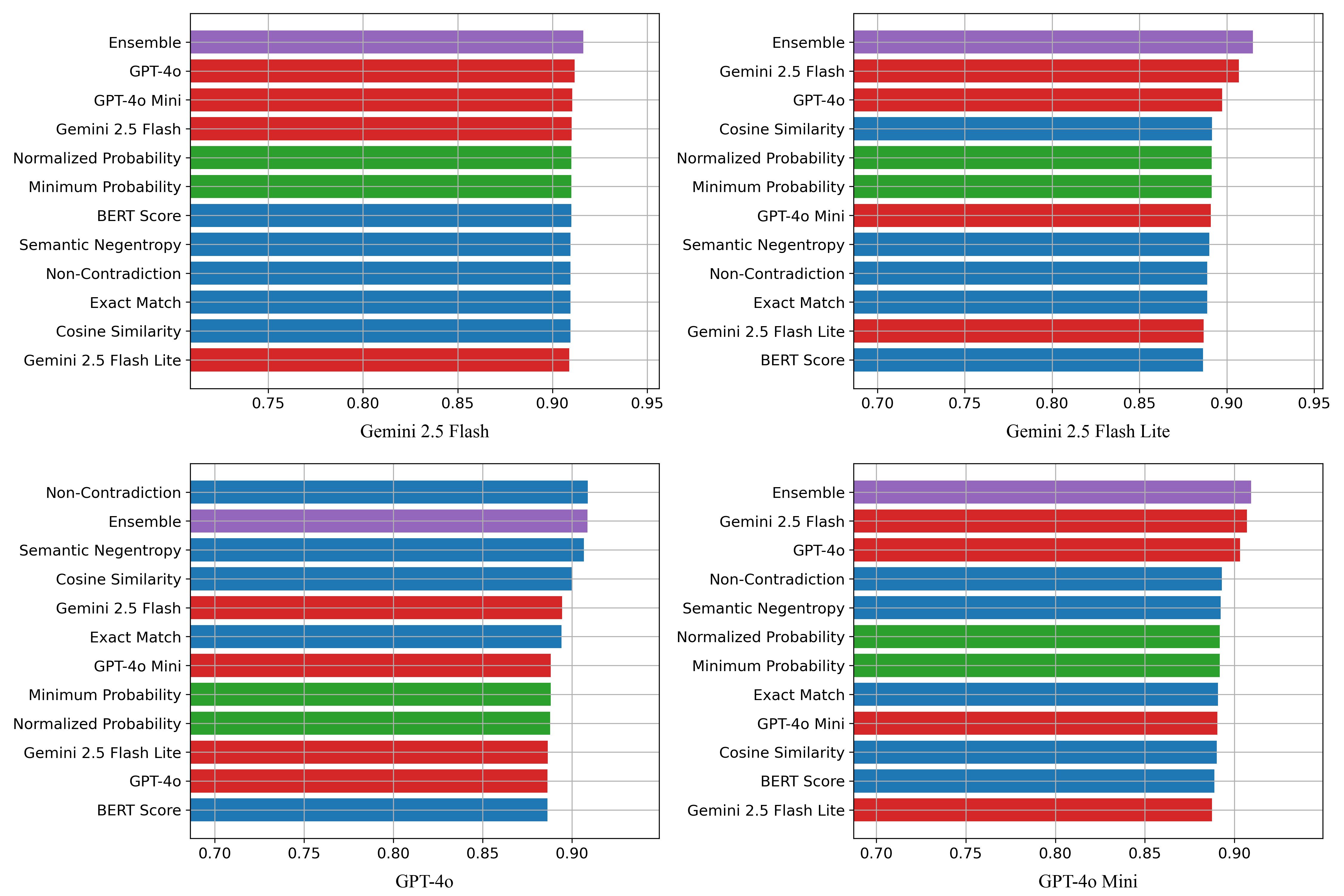}
        \caption{CSQA}
        \label{fig:e}
    \end{subfigure}
    \hfill
    \begin{subfigure}[b]{0.48\textwidth}
        \centering
        \includegraphics[width=\textwidth]{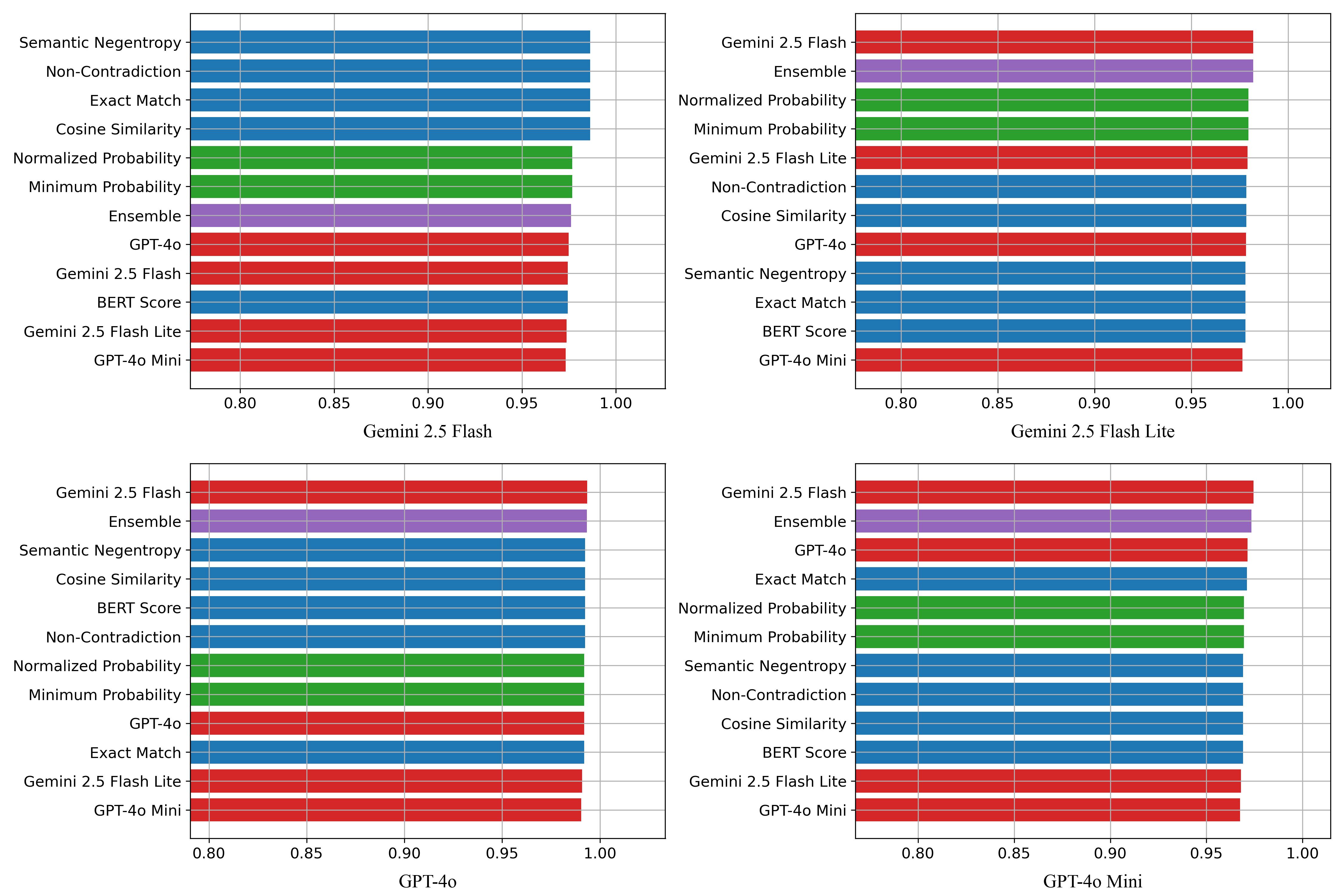}
        \caption{AI2-ARC}
        \label{fig:f}
    \end{subfigure}

    \caption{Scorer-Specific F1-Scores for Hallucination Detection by LLM and Dataset (Higher is Better)}
    \label{fig:f1}
\end{figure}

\begin{figure}[H]
    \centering
    \includegraphics[width=\textwidth]{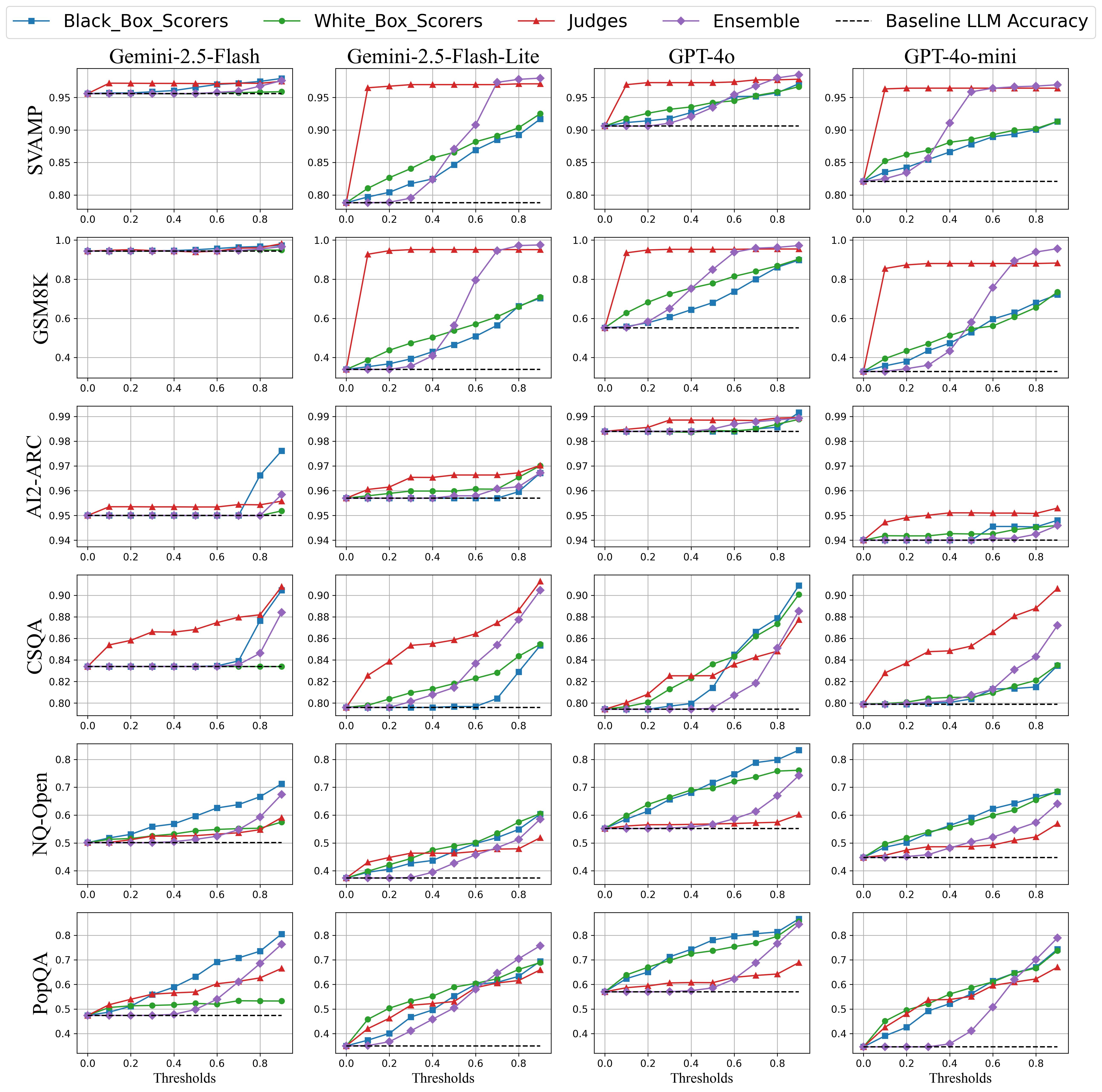}
    \caption{Filtered LLM Accuracy vs. Confidence Threshold (Top per Scorer Type)}
    \label{fig:acc}
\end{figure}

Figure \ref{fig:f1} displays the F1-scores for each scorer, while Table \ref{table:F1} summarizes the precision, recall, and F1 metrics for the top-performing scorer across all 24 evaluation scenarios. The results are largely consistent with the AUROC experiments. The ensemble scorer outperforms its individual components in most scenarios, achieving highest F1-score in 17 out of 24 scenarios. Similar to the AUROC experiments, the NLI-based scorers (NSN and NCP) often outperform other black-box scorers (18 out of 24 scenarios), consistent with the findings of previous studies \citep{kuhn2023semanticuncertaintylinguisticinvariances, manakul2023selfcheckgptzeroresourceblackboxhallucination, lin2024generatingconfidenceuncertaintyquantification, Farquhar2024}. Once again, the Gemini-2.5-Flash judge outperforms other LLM judges on the math benchmarks (8 out of 8 scenarios), while the GPT-4o judge outperforms other LLM judges on the short answer benchmarks (6 out of 8 scenarios). Interestingly, the strong LLM judge performance of Gemini-2.5-Flash on the math benchmarks and GPT-4o on the short answer benchmarks are consistent with their respective strengths when generating answers as the original LLM, where they achieved highest accuracy among the four LLMs on those same benchmarks (see Figure \ref{fig:acc}). Lastly, the two white-box scorers perform approximately equally, as in the AUROC experiments.


\paragraph{Filtered Accuracy@$\tau$.}
Lastly, we compute model accuracy on the subset of LLM responses having confidence scores exceeding a specified threshold $\tau$. We refer to this metric as \textit{Filtered Accuracy@}$\tau$. Since the LLM accuracy depends on the choice of the threshold $\tau$, we repeat the calculation for $\tau = 0, 0.1,..., 0.9$. Note that accuracy at $\tau=0$ uses the full sample without score-based filtering. 

Figure \ref{fig:acc} presents the Filtered Accuracy@$\tau$ for the highest performing white-box, black-box, LLM-as-a-Judge, and ensemble scorers. Across all scenarios, response filtering with the highest-performing scorers leads to an approximately monotonic increase in LLM accuracy as the threshold increases. For example, when filtering Gemini-2.5-Flash-Lite responses to PopQA questions using the leading white-box scorer, accuracy improves dramatically from a baseline of 0.35 to 0.61 at $\tau=0.6$. Similarly, with GPT-4o responses on GSM8K, filtering with the ensemble scorer achieves 0.93 accuracy at $\tau=0.6$, substantially higher than the baseline accuracy of 0.55.


\section{Discussion}
\paragraph{Choosing Among Scorers.} 
Choosing the right confidence scorer for an LLM system depends on several factors, including API support, latency requirements, LLM behavior, and the availability of graded datasets. If the API supports access to token probabilities in LLM generations, white-box scorers can be implemented without adding latency or generation costs.\footnote{The white-box scorers we consider in this work require only a single generation per prompt. However, sampling-based white-box methods also exist. See, for example, \cite{kuhn2023semanticuncertaintylinguisticinvariances, scalena2025eagerentropyawaregenerationadaptive, vashurin2025uncertaintyquantificationllmsminimum, Vashurin_2025, qiu2024semanticdensityuncertaintyquantification}.} However, if the API does not provide access to token probabilities, black-box scorers and LLM-as-a-Judge may be the only feasible options. When choosing among black-box and LLM-as-a-Judge scorers, latency requirements are a key consideration. For low-latency applications, practitioners should avoid higher-latency black-box scorers such as NLI-based scorers (NSN and NCP), opting instead for faster black-box scorers or LLM-as-a-Judge. If latency is not a concern, any of the black-box scorers may be suitable.


For LLM-as-a-Judge implementations, our findings reveal that an LLM's accuracy on a specific dataset positively relates to its ability to judge responses to questions from that same dataset, providing a practical criterion for judge selection. Finally, if a graded dataset is available, practitioners can tune an ensemble of various confidence scores to improve hallucination detection, as detailed in Appendix \ref{sec:tuning}. Our experiments demonstrate that a tuned ensemble can potentially provide more accurate confidence scores than individual scorers. By considering these factors and choosing the right confidence score, practitioners can improve the performance of their LLM system.

\paragraph{Using Confidence Scores.} In practice, practitioners may wish to use confidence scores for various purposes. First, practitioners can use our confidence scores for response filtering, where responses with low confidence are blocked, or `targeted' human-in-the-loop, where low-confidence responses are selected for manual review. Our experimental evaluations of Filtered Accuracy@$\tau$ demonstrate the efficacy of these approaches, illustrating notable improvements in LLM accuracy when low-confidence responses are filtered out. Note that selecting a confidence threshold for flagging or blocking responses will depend on the scorer used, the dataset being evaluated, and stakeholder values (e.g., relative cost of false negatives vs. false positives). 

Alternatively, confidence scores can be used for pre-deployment diagnostics, providing practitioners insights into the types of questions on which their LLM is performing worst. Findings from this type of exploratory analysis can inform strategies for improvements, such as further prompt engineering. Overall, the scorers included in our framework and toolkit provide practitioners with an actionable way to improve response quality, optimize resource allocation, and mitigate risks. 

We note two important ethical considerations related to using confidence scores. First, a confidence score reflects model uncertainty rather than ground-truth-based correctness, and high scores can induce over-reliance in high-risk settings. We therefore caution against using confidence scores as decision guarantees in domains such as medicine, law, or finance without external human review and domain-specific safeguards. Second, we encourage monitoring for distributional fairness and content diversity, including per-segment error-rate evaluation, and, where applicable, safeguards that prevent systematic removal of accurate responses for particular groups.

\paragraph{Limitations and Future Work.}
We note a few important limitations of this work. First, although our experiments leverage six question-answering benchmark datasets spanning three types of questions, we acknowledge that our findings on scorer-specific performance may not generalize to other types of questions.\footnote{For practical reasons, we selected benchmark datasets containing questions that could be graded computationally.} For example, long-form generation (such as summarization) can mix true and false claims within a single output and may require claim-level decomposition to obtain useful uncertainty signals. Likewise, correctness in code generation depends on syntax and execution, so black-box consistency measurements will likely require alternative methods.

Second, while we conduct experiments using four LLMs, performance of the various scorers may differ for other LLMs. Note that for different LLMs, differences in token probability distributions will impact the behavior of white-box scorers, and the degree of variation in responses to the same prompt will affect the performance of black-box scorers. LLM-as-a-Judge performance may vary significantly depending on the choice of LLM and instruction prompts used.

Additional limitations relate to the ensemble approach. First, while the ensemble approach is designed to optimize in-domain performance, we have not evaluated out-of-distribution generalization, including cross-dataset transfer of learned weights; we leave this to future work. Second, our experiments consider only linear ensembles. Future work should examine non-linear ensembling and its impact on hallucination detection, e.g., monotonic generalized additive models that retain interpretability while capturing modest non-linear effects, mixture-of-experts, and tree-based ensembles. These may outperform linear averaging when scorer signals interact or exhibit non-linear relationships. Practical trade-offs to study include latency, overfitting risk on small graded sets, and the need for interpretability in high-stakes deployments. Third, tuning the ensemble requires a graded dataset. For tasks with trivial grading (e.g., arithmetic, multiple-choice, short-answer), where there are well-defined, easy-to-recognize correct answers and comparisons with generated text can be computed automatically, this can be done by sampling a representative set of questions, generating responses, computing the selected UQ scores, and using the automatically derived labels to fit the weights. For tasks with non-trivial grading (e.g., summarization), practitioners can begin with a small human-graded dataset (e.g., a few hundred items); additional labels can then be added incrementally from production logs to further optimize the weights and improve ensemble performance.  

\section{Conclusions}
In this paper, we detail a framework for closed-book hallucination detection comprised of various black-box UQ, white-box UQ, and LLM-as-a-Judge scorers. To ensure standardized outputs of the scorers, we transform and normalize scorers (if necessary) such that all outputs range from 0 to 1, with higher scores indicating greater confidence in an LLM response. These response-level confidence scores can be used for generation-time hallucination detection across a wide variety of LLM use cases. Additionally, we introduce a novel, ensemble-based approach that leverages an optimized weighted average of any combination of individual confidence scores. Importantly, the extensible nature of our ensemble means that practitioners can include additional scorers as new methods become available. 

Our experimental evaluation of UQ-based scorers offers clear guidance for practitioners. Ensemble approaches consistently outperform individual methods for hallucination detection, with our findings strongly supporting use-case-specific customization rather than one-size-fits-all solutions. For those without token-probability access, NLI-based approaches typically provide the best black-box performance. Importantly, gains from sampling additional responses diminish as the number of candidate responses increases, offering a practical deployment guideline that balances effectiveness with computational efficiency. Finally, our analysis revealed that a model's accuracy on a specific dataset positively relates to its ability to judge responses to questions from that same dataset, providing a practical criterion for judge selection in evaluation frameworks.

The \texttt{uqlm} library implements all scorers presented and evaluated in this work. The repository is actively maintained. We welcome issues and pull requests and will continue to update integrations, add new scorers as research advances, and refresh examples as model and provider APIs change. 

\section*{Acknowledgements}
We wish to thank David Skarbrevik, Piero Ferrante, Xue (Crystal) Gu, Blake Aber, Viren Bajaj, Ho-Kyeong Ra, Zeya Ahmad,  Matthew Churgin,  Saicharan Sirangi, and Erik Widman for their helpful suggestions.

\section*{Conflict of Interest}
DB and MSC are employed and receive stock and equity from CVS Health® Corporation.

\bibliography{refs.bib}

@misc{scalena2025eagerentropyawaregenerationadaptive,
      title={EAGER: Entropy-Aware GEneRation for Adaptive Inference-Time Scaling}, 
      author={Daniel Scalena and Leonidas Zotos and Elisabetta Fersini and Malvina Nissim and Ahmet Üstün},
      year={2025},
      eprint={2510.11170},
      archivePrefix={arXiv},
      primaryClass={cs.LG},
      url={https://arxiv.org/abs/2510.11170}, 
}

@article{Vashurin_2025,
   title={Benchmarking Uncertainty Quantification Methods for Large Language Models with LM-Polygraph},
   volume={13},
   ISSN={2307-387X},
   url={http://dx.doi.org/10.1162/tacl_a_00737},
   DOI={10.1162/tacl_a_00737},
   journal={Transactions of the Association for Computational Linguistics},
   publisher={MIT Press},
   author={Vashurin, Roman and Fadeeva, Ekaterina and Vazhentsev, Artem and Rvanova, Lyudmila and Vasilev, Daniil and Tsvigun, Akim and Petrakov, Sergey and Xing, Rui and Sadallah, Abdelrahman and Grishchenkov, Kirill and Panchenko, Alexander and Baldwin, Timothy and Nakov, Preslav and Panov, Maxim and Shelmanov, Artem},
   year={2025},
   pages={220–248} }

@misc{vashurin2025uncertaintyquantificationllmsminimum,
      title={Uncertainty Quantification for LLMs through Minimum Bayes Risk: Bridging Confidence and Consistency}, 
      author={Roman Vashurin and Maiya Goloburda and Albina Ilina and Aleksandr Rubashevskii and Preslav Nakov and Artem Shelmanov and Maxim Panov},
      year={2025},
      eprint={2502.04964},
      archivePrefix={arXiv},
      primaryClass={cs.CL},
      url={https://arxiv.org/abs/2502.04964}, 
}

@inproceedings{Akiba_Optuna_A_next-generation_2019,
author = {Akiba, Takuya and Sano, Shotaro and Yanase, Toshihiko and Ohta, Takeru and Koyama, Masanori},
booktitle = {Proceedings of the 25th ACM SIGKDD international conference on knowledge discovery \& data mining},
doi = {10.1145/3292500.3330701},
pages = {2623--2631},
title = {{Optuna: A next-generation hyperparameter optimization framework}},
year = {2019}
}

@misc{OpenAI_doc, 
url={https://platform.openai.com/docs/models}, journal={OpenAI.com}, author={OpenAI}}

@misc{gemini_doc, 
url={https://cloud.google.com/vertex-ai/generative-ai/docs/models}, journal={Google Cloud}, author={Google}}

@misc{bouchard2025uqlmpythonpackageuncertainty,
      title={UQLM: A Python Package for Uncertainty Quantification in Large Language Models}, 
      author={Dylan Bouchard and Mohit Singh Chauhan and David Skarbrevik and Ho-Kyeong Ra and Viren Bajaj and Zeya Ahmad},
      year={2025},
      eprint={2507.06196},
      archivePrefix={arXiv},
      primaryClass={cs.CL},
      url={https://arxiv.org/abs/2507.06196}, 
}

@misc{OpenAI, 
title={Introducing GPT-4.5 | OpenAI}, 
url={https://openai.com/index/introducing-gpt-4-5/}, 
journal={OpenAI.com}, 
author={OpenAI},
year={2025}}

@misc{jiang2024graphbaseduncertaintymetricslongform,
      title={Graph-based Uncertainty Metrics for Long-form Language Model Outputs}, 
      author={Mingjian Jiang and Yangjun Ruan and Prasanna Sattigeri and Salim Roukos and Tatsunori Hashimoto},
      year={2024},
      eprint={2410.20783},
      archivePrefix={arXiv},
      primaryClass={cs.CL},
      url={https://arxiv.org/abs/2410.20783}, 
}

@misc{wang2024decodingtrustcomprehensiveassessmenttrustworthiness,
      title={DecodingTrust: A Comprehensive Assessment of Trustworthiness in GPT Models}, 
      author={Boxin Wang and Weixin Chen and Hengzhi Pei and Chulin Xie and Mintong Kang and Chenhui Zhang and Chejian Xu and Zidi Xiong and Ritik Dutta and Rylan Schaeffer and Sang T. Truong and Simran Arora and Mantas Mazeika and Dan Hendrycks and Zinan Lin and Yu Cheng and Sanmi Koyejo and Dawn Song and Bo Li},
      year={2024},
      eprint={2306.11698},
      archivePrefix={arXiv},
      primaryClass={cs.CL},
      url={https://arxiv.org/abs/2306.11698}, 
}

@misc{huang2024surveyuncertaintyestimationllms,
      title={A Survey of Uncertainty Estimation in LLMs: Theory Meets Practice}, 
      author={Hsiu-Yuan Huang and Yutong Yang and Zhaoxi Zhang and Sanwoo Lee and Yunfang Wu},
      year={2024},
      eprint={2410.15326},
      archivePrefix={arXiv},
      primaryClass={cs.CL},
      url={https://arxiv.org/abs/2410.15326}, 
}

@misc{shorinwa2024surveyuncertaintyquantificationlarge,
      title={A Survey on Uncertainty Quantification of Large Language Models: Taxonomy, Open Research Challenges, and Future Directions}, 
      author={Ola Shorinwa and Zhiting Mei and Justin Lidard and Allen Z. Ren and Anirudha Majumdar},
      year={2024},
      eprint={2412.05563},
      archivePrefix={arXiv},
      primaryClass={cs.CL},
      url={https://arxiv.org/abs/2412.05563}, 
}

@misc{huang2023surveyhallucinationlargelanguage,
      title={A Survey on Hallucination in Large Language Models: Principles, Taxonomy, Challenges, and Open Questions}, 
      author={Lei Huang and Weijiang Yu and Weitao Ma and Weihong Zhong and Zhangyin Feng and Haotian Wang and Qianglong Chen and Weihua Peng and Xiaocheng Feng and Bing Qin and Ting Liu},
      year={2023},
      eprint={2311.05232},
      archivePrefix={arXiv},
      primaryClass={cs.CL},
      url={https://arxiv.org/abs/2311.05232}, 
}

@misc{tonmoy2024comprehensivesurveyhallucinationmitigation,
      title={A Comprehensive Survey of Hallucination Mitigation Techniques in Large Language Models}, 
      author={S. M Towhidul Islam Tonmoy and S M Mehedi Zaman and Vinija Jain and Anku Rani and Vipula Rawte and Aman Chadha and Amitava Das},
      year={2024},
      eprint={2401.01313},
      archivePrefix={arXiv},
      primaryClass={cs.CL},
      url={https://arxiv.org/abs/2401.01313}, 
}

@misc{chen2023quantifyinguncertaintyanswerslanguage,
      title={Quantifying Uncertainty in Answers from any Language Model and Enhancing their Trustworthiness}, 
      author={Jiuhai Chen and Jonas Mueller},
      year={2023},
      eprint={2308.16175},
      archivePrefix={arXiv},
      primaryClass={cs.CL},
      url={https://arxiv.org/abs/2308.16175}, 
}

@Article{Farquhar2024,
author={Farquhar, Sebastian
and Kossen, Jannik
and Kuhn, Lorenz
and Gal, Yarin},
title={Detecting hallucinations in large language models using semantic entropy},
journal={Nature},
year={2024},
month={Jun},
day={01},
volume={630},
number={8017},
pages={625-630},
issn={1476-4687},
doi={10.1038/s41586-024-07421-0},
url={https://doi.org/10.1038/s41586-024-07421-0}
}

@misc{zhang2024luqlongtextuncertaintyquantification,
      title={LUQ: Long-text Uncertainty Quantification for LLMs}, 
      author={Caiqi Zhang and Fangyu Liu and Marco Basaldella and Nigel Collier},
      year={2024},
      eprint={2403.20279},
      archivePrefix={arXiv},
      primaryClass={cs.CL},
      url={https://arxiv.org/abs/2403.20279}, 
}

@misc{manakul2023selfcheckgptzeroresourceblackboxhallucination,
      title={SelfCheckGPT: Zero-Resource Black-Box Hallucination Detection for Generative Large Language Models}, 
      author={Potsawee Manakul and Adian Liusie and Mark J. F. Gales},
      year={2023},
      eprint={2303.08896},
      archivePrefix={arXiv},
      primaryClass={cs.CL},
      url={https://arxiv.org/abs/2303.08896}, 
}

@misc{kossen2024semanticentropyprobesrobust,
      title={Semantic Entropy Probes: Robust and Cheap Hallucination Detection in LLMs}, 
      author={Jannik Kossen and Jiatong Han and Muhammed Razzak and Lisa Schut and Shreshth Malik and Yarin Gal},
      year={2024},
      eprint={2406.15927},
      archivePrefix={arXiv},
      primaryClass={cs.CL},
      url={https://arxiv.org/abs/2406.15927}, 
}

@misc{lin2024generatingconfidenceuncertaintyquantification,
      title={Generating with Confidence: Uncertainty Quantification for Black-box Large Language Models}, 
      author={Zhen Lin and Shubhendu Trivedi and Jimeng Sun},
      year={2024},
      eprint={2305.19187},
      archivePrefix={arXiv},
      primaryClass={cs.CL},
      url={https://arxiv.org/abs/2305.19187}, 
}

@misc{kuhn2023semanticuncertaintylinguisticinvariances,
      title={Semantic Uncertainty: Linguistic Invariances for Uncertainty Estimation in Natural Language Generation}, 
      author={Lorenz Kuhn and Yarin Gal and Sebastian Farquhar},
      year={2023},
      eprint={2302.09664},
      archivePrefix={arXiv},
      primaryClass={cs.CL},
      url={https://arxiv.org/abs/2302.09664}, 
}

@misc{qiu2024semanticdensityuncertaintyquantification,
      title={Semantic Density: Uncertainty Quantification for Large Language Models through Confidence Measurement in Semantic Space}, 
      author={Xin Qiu and Risto Miikkulainen},
      year={2024},
      eprint={2405.13845},
      archivePrefix={arXiv},
      primaryClass={cs.CL},
      url={https://arxiv.org/abs/2405.13845}, 
}

@misc{cole2023selectivelyansweringambiguousquestions,
      title={Selectively Answering Ambiguous Questions}, 
      author={Jeremy R. Cole and Michael J. Q. Zhang and Daniel Gillick and Julian Martin Eisenschlos and Bhuwan Dhingra and Jacob Eisenstein},
      year={2023},
      eprint={2305.14613},
      archivePrefix={arXiv},
      primaryClass={cs.CL},
      url={https://arxiv.org/abs/2305.14613}, 
}

@inproceedings{lin-2004-rouge,
    title = "{ROUGE}: A Package for Automatic Evaluation of Summaries",
    author = "Lin, Chin-Yew",
    booktitle = "Text Summarization Branches Out",
    month = jul,
    year = "2004",
    address = "Barcelona, Spain",
    publisher = "Association for Computational Linguistics",
    url = "https://aclanthology.org/W04-1013/",
    pages = "74--81"
}

@inproceedings{10.3115/1073083.1073135,
author = {Papineni, Kishore and Roukos, Salim and Ward, Todd and Zhu, Wei-Jing},
title = {BLEU: a method for automatic evaluation of machine translation},
year = {2002},
publisher = {Association for Computational Linguistics},
address = {USA},
url = {https://doi.org/10.3115/1073083.1073135},
doi = {10.3115/1073083.1073135},
booktitle = {Proceedings of the 40th Annual Meeting on Association for Computational Linguistics},
pages = {311–318},
numpages = {8},
location = {Philadelphia, Pennsylvania},
series = {ACL '02}
}

@INPROCEEDINGS{9194665,
  author={Qurashi, Abdul Wahab and Holmes, Violeta and Johnson, Anju P.},
  booktitle={2020 International Conference on INnovations in Intelligent SysTems and Applications (INISTA)}, 
  title={Document Processing: Methods for Semantic Text Similarity Analysis}, 
  year={2020},
  volume={},
  number={},
  pages={1-6},
doi={10.1109/INISTA49547.2020.9194665}}

@inproceedings{banerjee-lavie-2005-meteor,
    title = "{METEOR}: An Automatic Metric for {MT} Evaluation with Improved Correlation with Human Judgments",
    author = "Banerjee, Satanjeev  and
      Lavie, Alon",
    editor = "Goldstein, Jade  and
      Lavie, Alon  and
      Lin, Chin-Yew  and
      Voss, Clare",
    booktitle = "Proceedings of the {ACL} Workshop on Intrinsic and Extrinsic Evaluation Measures for Machine Translation and/or Summarization",
    month = jun,
    year = "2005",
    address = "Ann Arbor, Michigan",
    publisher = "Association for Computational Linguistics",
    url = "https://aclanthology.org/W05-0909/",
    pages = "65--72"
}

@misc{reimers2019sentencebertsentenceembeddingsusing,
      title={Sentence-BERT: Sentence Embeddings using Siamese BERT-Networks}, 
      author={Nils Reimers and Iryna Gurevych},
      year={2019},
      eprint={1908.10084},
      archivePrefix={arXiv},
      primaryClass={cs.CL},
      url={https://arxiv.org/abs/1908.10084}, 
}

@misc{zhang2020bertscoreevaluatingtextgeneration,
      title={BERTScore: Evaluating Text Generation with BERT}, 
      author={Tianyi Zhang and Varsha Kishore and Felix Wu and Kilian Q. Weinberger and Yoav Artzi},
      year={2020},
      eprint={1904.09675},
      archivePrefix={arXiv},
      primaryClass={cs.CL},
      url={https://arxiv.org/abs/1904.09675}, 
}

@misc{agrawal2024languagemodelsknowtheyre,
      title={Do Language Models Know When They're Hallucinating References?}, 
      author={Ayush Agrawal and Mirac Suzgun and Lester Mackey and Adam Tauman Kalai},
      year={2024},
      eprint={2305.18248},
      archivePrefix={arXiv},
      primaryClass={cs.CL},
      url={https://arxiv.org/abs/2305.18248}, 
}

@misc{xiong2024llmsexpressuncertaintyempirical,
      title={Can LLMs Express Their Uncertainty? An Empirical Evaluation of Confidence Elicitation in LLMs}, 
      author={Miao Xiong and Zhiyuan Hu and Xinyang Lu and Yifei Li and Jie Fu and Junxian He and Bryan Hooi},
      year={2024},
      eprint={2306.13063},
      archivePrefix={arXiv},
      primaryClass={cs.CL},
      url={https://arxiv.org/abs/2306.13063}, 
}

@misc{kadavath2022languagemodelsmostlyknow,
      title={Language Models (Mostly) Know What They Know}, 
      author={Saurav Kadavath and Tom Conerly and Amanda Askell and Tom Henighan and Dawn Drain and Ethan Perez and Nicholas Schiefer and Zac Hatfield-Dodds and Nova DasSarma and Eli Tran-Johnson and Scott Johnston and Sheer El-Showk and Andy Jones and Nelson Elhage and Tristan Hume and Anna Chen and Yuntao Bai and Sam Bowman and Stanislav Fort and Deep Ganguli and Danny Hernandez and Josh Jacobson and Jackson Kernion and Shauna Kravec and Liane Lovitt and Kamal Ndousse and Catherine Olsson and Sam Ringer and Dario Amodei and Tom Brown and Jack Clark and Nicholas Joseph and Ben Mann and Sam McCandlish and Chris Olah and Jared Kaplan},
      year={2022},
      eprint={2207.05221},
      archivePrefix={arXiv},
      primaryClass={cs.CL},
      url={https://arxiv.org/abs/2207.05221}, 
}

@misc{cohen2023lmvslmdetecting,
      title={LM vs LM: Detecting Factual Errors via Cross Examination}, 
      author={Roi Cohen and May Hamri and Mor Geva and Amir Globerson},
      year={2023},
      eprint={2305.13281},
      archivePrefix={arXiv},
      primaryClass={cs.CL},
      url={https://arxiv.org/abs/2305.13281}, 
}

@misc{verga2024replacingjudgesjuriesevaluating,
      title={Replacing Judges with Juries: Evaluating LLM Generations with a Panel of Diverse Models}, 
      author={Pat Verga and Sebastian Hofstatter and Sophia Althammer and Yixuan Su and Aleksandra Piktus and Arkady Arkhangorodsky and Minjie Xu and Naomi White and Patrick Lewis},
      year={2024},
      eprint={2404.18796},
      archivePrefix={arXiv},
      primaryClass={cs.CL},
      url={https://arxiv.org/abs/2404.18796}, 
}

@misc{ling2024uncertaintyquantificationincontextlearning,
      title={Uncertainty Quantification for In-Context Learning of Large Language Models}, 
      author={Chen Ling and Xujiang Zhao and Xuchao Zhang and Wei Cheng and Yanchi Liu and Yiyou Sun and Mika Oishi and Takao Osaki and Katsushi Matsuda and Jie Ji and Guangji Bai and Liang Zhao and Haifeng Chen},
      year={2024},
      eprint={2402.10189},
      archivePrefix={arXiv},
      primaryClass={cs.CL},
      url={https://arxiv.org/abs/2402.10189}, 
}

@misc{bakman2024marsmeaningawareresponsescoring,
      title={MARS: Meaning-Aware Response Scoring for Uncertainty Estimation in Generative LLMs}, 
      author={Yavuz Faruk Bakman and Duygu Nur Yaldiz and Baturalp Buyukates and Chenyang Tao and Dimitrios Dimitriadis and Salman Avestimehr},
      year={2024},
      eprint={2402.11756},
      archivePrefix={arXiv},
      primaryClass={cs.CL},
      url={https://arxiv.org/abs/2402.11756}, 
}

@misc{fadeeva2024factcheckingoutputlargelanguage,
      title={Fact-Checking the Output of Large Language Models via Token-Level Uncertainty Quantification}, 
      author={Ekaterina Fadeeva and Aleksandr Rubashevskii and Artem Shelmanov and Sergey Petrakov and Haonan Li and Hamdy Mubarak and Evgenii Tsymbalov and Gleb Kuzmin and Alexander Panchenko and Timothy Baldwin and Preslav Nakov and Maxim Panov},
      year={2024},
      eprint={2403.04696},
      archivePrefix={arXiv},
      primaryClass={cs.CL},
      url={https://arxiv.org/abs/2403.04696}, 
}

@misc{guerreiro2023lookingneedlehaystackcomprehensive,
      title={Looking for a Needle in a Haystack: A Comprehensive Study of Hallucinations in Neural Machine Translation}, 
      author={Nuno M. Guerreiro and Elena Voita and André F. T. Martins},
      year={2023},
      eprint={2208.05309},
      archivePrefix={arXiv},
      primaryClass={cs.CL},
      url={https://arxiv.org/abs/2208.05309}, 
}

@misc{zhang2023enhancinguncertaintybasedhallucinationdetection,
      title={Enhancing Uncertainty-Based Hallucination Detection with Stronger Focus}, 
      author={Tianhang Zhang and Lin Qiu and Qipeng Guo and Cheng Deng and Yue Zhang and Zheng Zhang and Chenghu Zhou and Xinbing Wang and Luoyi Fu},
      year={2023},
      eprint={2311.13230},
      archivePrefix={arXiv},
      primaryClass={cs.CL},
      url={https://arxiv.org/abs/2311.13230}, 
}

@misc{varshney2023stitchtimesavesnine,
      title={A Stitch in Time Saves Nine: Detecting and Mitigating Hallucinations of LLMs by Validating Low-Confidence Generation}, 
      author={Neeraj Varshney and Wenlin Yao and Hongming Zhang and Jianshu Chen and Dong Yu},
      year={2023},
      eprint={2307.03987},
      archivePrefix={arXiv},
      primaryClass={cs.CL},
      url={https://arxiv.org/abs/2307.03987}, 
}

@misc{luo2023zeroresourcehallucinationpreventionlarge,
      title={Zero-Resource Hallucination Prevention for Large Language Models}, 
      author={Junyu Luo and Cao Xiao and Fenglong Ma},
      year={2023},
      eprint={2309.02654},
      archivePrefix={arXiv},
      primaryClass={cs.CL},
      url={https://arxiv.org/abs/2309.02654}, 
}

@misc{ren2023selfevaluationimprovesselectivegeneration,
      title={Self-Evaluation Improves Selective Generation in Large Language Models}, 
      author={Jie Ren and Yao Zhao and Tu Vu and Peter J. Liu and Balaji Lakshminarayanan},
      year={2023},
      eprint={2312.09300},
      archivePrefix={arXiv},
      primaryClass={cs.CL},
      url={https://arxiv.org/abs/2312.09300}, 
}

@misc{vanderpoel2022mutualinformationalleviateshallucinations,
      title={Mutual Information Alleviates Hallucinations in Abstractive Summarization}, 
      author={Liam van der Poel and Ryan Cotterell and Clara Meister},
      year={2022},
      eprint={2210.13210},
      archivePrefix={arXiv},
      primaryClass={cs.CL},
      url={https://arxiv.org/abs/2210.13210}, 
}

@misc{wang2023selfconsistencyimproveschainthought,
      title={Self-Consistency Improves Chain of Thought Reasoning in Language Models}, 
      author={Xuezhi Wang and Jason Wei and Dale Schuurmans and Quoc Le and Ed Chi and Sharan Narang and Aakanksha Chowdhery and Denny Zhou},
      year={2023},
      eprint={2203.11171},
      archivePrefix={arXiv},
      primaryClass={cs.CL},
      url={https://arxiv.org/abs/2203.11171}, 
}

@misc{malinin2021uncertaintyestimationautoregressivestructured,
      title={Uncertainty Estimation in Autoregressive Structured Prediction}, 
      author={Andrey Malinin and Mark Gales},
      year={2021},
      eprint={2002.07650},
      archivePrefix={arXiv},
      primaryClass={stat.ML},
      url={https://arxiv.org/abs/2002.07650}, 
}

@misc{cobbe2021trainingverifierssolvemath,
      title={Training Verifiers to Solve Math Word Problems}, 
      author={Karl Cobbe and Vineet Kosaraju and Mohammad Bavarian and Mark Chen and Heewoo Jun and Lukasz Kaiser and Matthias Plappert and Jerry Tworek and Jacob Hilton and Reiichiro Nakano and Christopher Hesse and John Schulman},
      year={2021},
      eprint={2110.14168},
      archivePrefix={arXiv},
      primaryClass={cs.LG},
      url={https://arxiv.org/abs/2110.14168}, 
}

@misc{patel2021nlpmodelsreallyable,
      title={Are NLP Models really able to Solve Simple Math Word Problems?}, 
      author={Arkil Patel and Satwik Bhattamishra and Navin Goyal},
      year={2021},
      eprint={2103.07191},
      archivePrefix={arXiv},
      primaryClass={cs.CL},
      url={https://arxiv.org/abs/2103.07191}, 
}

@inproceedings{mallen-etal-2023-trust,
    title = "When Not to Trust Language Models: Investigating Effectiveness of Parametric and Non-Parametric Memories",
    author = "Mallen, Alex  and
      Asai, Akari  and
      Zhong, Victor  and
      Das, Rajarshi  and
      Khashabi, Daniel  and
      Hajishirzi, Hannaneh",
    editor = "Rogers, Anna  and
      Boyd-Graber, Jordan  and
      Okazaki, Naoaki",
    booktitle = "Proceedings of the 61st Annual Meeting of the Association for Computational Linguistics (Volume 1: Long Papers)",
    month = jul,
    year = "2023",
    address = "Toronto, Canada",
    publisher = "Association for Computational Linguistics",
    url = "https://aclanthology.org/2023.acl-long.546/",
    doi = "10.18653/v1/2023.acl-long.546",
    pages = "9802--9822"
}

@misc{lee2019latentretrievalweaklysupervised,
      title={Latent Retrieval for Weakly Supervised Open Domain Question Answering}, 
      author={Kenton Lee and Ming-Wei Chang and Kristina Toutanova},
      year={2019},
      eprint={1906.00300},
      archivePrefix={arXiv},
      primaryClass={cs.CL},
      url={https://arxiv.org/abs/1906.00300}, 
}

@misc{clark2018thinksolvedquestionanswering,
      title={Think you have Solved Question Answering? Try ARC, the AI2 Reasoning Challenge}, 
      author={Peter Clark and Isaac Cowhey and Oren Etzioni and Tushar Khot and Ashish Sabharwal and Carissa Schoenick and Oyvind Tafjord},
      year={2018},
      eprint={1803.05457},
      archivePrefix={arXiv},
      primaryClass={cs.AI},
      url={https://arxiv.org/abs/1803.05457}, 
}

@misc{talmor2022commonsenseqa20exposinglimits,
      title={CommonsenseQA 2.0: Exposing the Limits of AI through Gamification}, 
      author={Alon Talmor and Ori Yoran and Ronan Le Bras and Chandra Bhagavatula and Yoav Goldberg and Yejin Choi and Jonathan Berant},
      year={2022},
      eprint={2201.05320},
      archivePrefix={arXiv},
      primaryClass={cs.CL},
      url={https://arxiv.org/abs/2201.05320}, 
}
\bibliographystyle{tmlr}

\newpage
\appendix

\section{LLM-as-a-Judge Prompt}
\label{sec:judge_prompt}
Our LLM-as-a-Judge scorer used the following instruction prompt: 

\begin{quote}
Question: [question], Proposed Answer: [answer]. \\
\\
How likely is the above answer to be correct? Analyze the answer and give your confidence in this answer between 0 (lowest) and 100 (highest), with 100 being certain the answer is correct, and 0 being certain the answer is incorrect. THE CONFIDENCE RATING YOU PROVIDE MUST BE BETWEEN 0 and 100. ONLY RETURN YOUR NUMERICAL SCORE WITH NO SURROUNDING TEXT OR EXPLANATION. \\
\\
\# Example 1 \\
\#\# Data to analyze \\
Question: Who was the first president of the United States?, Proposed Answer: Benjamin Franklin. \\
\\
\#\# Your response \\
4 (highly certain the proposed answer is incorrect) \\ 
\\

\# Example 2 \\
\#\# Data to analyze \\
Question: What is 2+2?, Proposed Answer: 4 \\
\\
\#\# Your response \\
99 (highly certain the proposed answer is correct) \\
\end{quote}
To ensure a normalized confidence score consistent with the other scorers, we normalize the value returned by the LLM judge to be between 0 and 1. The capitalization and repeated instructions, inspired by \citet{wang2024decodingtrustcomprehensiveassessmenttrustworthiness}, are included to ensure the LLM correctly follows instructions.

\section{Ensemble Tuning}
\label{sec:tuning}

We outline a method for tuning ensemble weights for improved hallucination detection accuracy. This approach allows for customizable component-importance that can be optimized for a specific use case. In practice, tuning the ensemble weights requires having a `graded' set of $n$ original LLM responses which indicate whether a hallucination is present in each response.\footnote{Grading responses may be accomplished computationally for certain tasks, e.g. multiple choice questions. However, in many cases, this will require a human grader to manually evaluate the set of responses.} Given a set of $n$ prompts, denoted as $\mathbf{x}$:
\begin{equation}
\mathbf{x} = \begin{pmatrix}
x_1  \\
x_2 \\
\vdots  \\
x_n  \\
\end{pmatrix},
\end{equation}

denote corresponding correct reference answers as  $\mathbf{y}^*$,
\begin{equation}
\mathbf{y}^* = \begin{pmatrix}
y_1^*  \\
y_2^* \\
\vdots  \\
y_n^*  \\
\end{pmatrix},
\end{equation}

original LLM responses as $\mathbf{y}$,
\begin{equation}
\mathbf{y} = \begin{pmatrix}
y_1  \\
y_2 \\
\vdots  \\
y_n  \\
\end{pmatrix},
\end{equation}
and candidate responses across all prompts with the matrix $\Tilde{\mathbf{Y} }$:
\begin{equation}
\Tilde{\mathbf{Y} }= 
\begin{pmatrix}
\Tilde{\mathbf{y}}_1  \\
\Tilde{\mathbf{y}}_2 \\
\vdots  \\
\Tilde{\mathbf{y}}_n  \\
\end{pmatrix}
=
\begin{pmatrix}
\Tilde{y}_{11} & \Tilde{y}_{12} & \cdots & \Tilde{y}_{1m} \\
\Tilde{y}_{21} & \Tilde{y}_{22} & \cdots & \Tilde{y}_{2m} \\
\vdots & \vdots & \ddots & \vdots \\
\Tilde{y}_{n1} & \Tilde{y}_{n2} & \cdots & \Tilde{y}_{nm} \\
\end{pmatrix}.
\end{equation}

Analogously, we denote the vectors of ensemble confidence scores, binary ensemble hallucination predictions, and corresponding ground truth values respectively as 
\begin{equation}
\hat{\mathbf{s}}(\mathbf{y}; \Tilde{\mathbf{Y}}, \mathbf{x}, \mathbf{w}) = \begin{pmatrix}
\hat{s}(y_1; \Tilde{\mathbf{y}}_1, x_1, \mathbf{w})  \\
\hat{s}(y_2; \Tilde{\mathbf{y}}_2, x_2, \mathbf{w}) \\
\vdots  \\
\hat{s}(y_n; \Tilde{\mathbf{y}}_n, x_n, \mathbf{w})  \\
\end{pmatrix},
\end{equation}
\begin{equation}
\hat{\mathbf{h}}(\mathbf{y}; \Tilde{\mathbf{Y}}, \mathbf{x}, \mathbf{w}, \tau) = \begin{pmatrix}
\hat{h}(y_1; \Tilde{\mathbf{y}}_1, x_1, \mathbf{w}, \tau)  \\
\hat{h}(y_2; \Tilde{\mathbf{y}}_2, x_2, \mathbf{w}, \tau) \\
\vdots  \\
\hat{h}(y_n; \Tilde{\mathbf{y}}_n, x_n, \mathbf{w}, \tau)  \\
\end{pmatrix},
\end{equation}
and
\begin{equation}
\mathbf{h}(\mathbf{y}; \mathbf{y}^*, \mathbf{x}) = \begin{pmatrix}
h(y_1; y_1^*, x_1)  \\
h(y_2; y_2^*, x_2) \\
\vdots  \\
h(y_n; y_n^*, x_n)  \\
\end{pmatrix},
\end{equation}

Modeling this problem as binary classification enables us to tune the weights of our ensemble classifier using standard classification objective functions. Following this approach, we consider two distinct strategies to tune ensemble weights $w_1,...,w_K$: threshold-agnostic optimization and threshold-aware optimization. 

\paragraph{Threshold-Agnostic Weights Optimization.}
Our first ensemble tuning strategy uses a threshold-agnostic objective function for tuning the ensemble weights. Given a set of $n$ prompts, corresponding original LLM responses and candidate responses, the optimal set of weights, $\mathbf{w}^*$, is the solution to the following problem:

\begin{equation}
\mathbf{w}^* = \argmax_{\mathbf{w} \in \mathcal{W}} \mathcal{S}( \hat{\mathbf{s}}(\mathbf{y}; \Tilde{\mathbf{Y}}, \mathbf{x}, \mathbf{w}), \mathbf{h}(\mathbf{y}; \mathbf{y}^*, \mathbf{x})),
\end{equation}
where

\begin{equation}
    \mathcal{W} = \{(w_1,...,w_K): \sum_{k=1}^K w_k =1, w_k \geq 0  \text{ } \forall \text{ }  k =1,...,K\}
\end{equation}
is the support of the ensemble weights and $\mathcal{S}$ is a threshold-agnostic classification performance metric, such as area under the receiver-operator characteristic curve (AUROC).

After optimizing the weights, we subsequently tune the threshold using a threshold-dependent objective function. Hence, the optimal threshold, $\tau^*$, is the solution to the following optimization problem:

\begin{equation}
\mathbf{\tau}^* = \argmax_{\tau \in (0,1)} \mathcal{B}(\hat{\mathbf{h}}(\mathbf{y}; \Tilde{\mathbf{Y}}, \mathbf{x}, \mathbf{w}^*, \tau ), {\mathbf{h}(\mathbf{y}; \mathbf{y}^*, \mathbf{x}))},
\end{equation}
where $\mathcal{B}$ is a threshold-dependent classification performance metric, such as F1-score.

\paragraph{Threshold-Aware Weights Optimization.}
Alternatively, practitioners may wish jointly optimize ensemble weights and classification threshold using the same objective. This type of optimization relies on a threshold-dependent objective. We can write this optimization problem as follows:

\begin{equation}
\mathbf{w}^*, \tau^* = \argmax_{\mathbf{w} \in \mathcal{W}, \tau \in (0,1)} \mathcal{B}(\hat{\mathbf{h}}(\mathbf{y}; \Tilde{\mathbf{Y}}, \mathbf{x}, \mathbf{w}, \tau ), {\mathbf{h}(\mathbf{y}; \mathbf{y}^*, \mathbf{x}))},
\end{equation}
where terms follow the same definitions as above.

\section{Response Grading}
\label{sec:grading}

 For each task-type, the LLM is instructed to output its response in a specific form.  We therefore instantiate $h(\cdot)$ with task-specific graders aligned to that form. Unless noted, grading is case-insensitive and trims whitespace.\footnote{These automatic graders may not be perfectly accurate in every instance; residual errors can arise when model outputs deviate from the specified response format. Their effectiveness therefore depends on instruction adherence, and they should be viewed as practical proxies for answer-key–based factual correctness.}

\paragraph{Math (numeric answer only).}
For math questions, the LLM is provided the following instruction: ``Return only the numerical answer with no additional text.''  
Let $y_i$ be the model output. Let $\text{int}(y_i)$ extract the leading integer substring from $y_i$ if present; otherwise return $\varnothing$. Let $y_i^* \in \mathbb{Z}$ be the correct integer answer.  
\[
h_{\text{math}}(y_i; y_i^*) =
\begin{cases}
0 & \text{if } \text{int}(y_i) = y_i^*,\\
1 & \text{otherwise.}
\end{cases}
\]
Non-numeric or missing integer outputs are graded as incorrect.

\paragraph{Multiple choice (letter only).}
For multiple choice questions, the LLM is provided the following instruction: ``Return only the letter of the response with no additional text or explanation.'' 
Let $\mathrm{norm}_{mc}(y_i) \in \{A,B,C,D,E\}$ denote response $y_i$ after normalization to uppercase and trimming. Let $y_i^* \in \{A,B,C,D,E\}$ be the correct letter.
\[
h_{\text{mc}}(y_i; y_i^*) =
\begin{cases}
0 & \text{if } \mathrm{norm}_{mc}(y_i) = y_i^*,\\
1 & \text{otherwise.}
\end{cases}
\]
Any invalid response based on the provided instruction, i.e. $\mathrm{norm}_{mc}(y_i) \notin \{A,B,C,D,E\}$, is graded as incorrect.

\paragraph{Short-answer (keyword match).}
Lastly, for short-answer questions, the LLM is instructed as follows: ``Return only the answer as concisely as possible without providing an explanation.''  
Let $\mathbf{y}_i^* = \{y_{i1}^*, \dots, y_{iA}^*\}$ be the set of acceptable answers.\footnote{The short-answer datasets contain multiple versions of the same correct answer to allow for phrasing flexibility. For example, acceptable answers to the question ``What is Tobias Lindholm's occupation?'' include \{``film director'', ``movie director'', ``director'', ``motion picture director'', ``screenwriter'', ``scenarist'', ``writer'', ``screen writer'', ``script writer'', ``scriptwriter''\}.} Define a normalization function $\mathrm{norm}_{sa}(y_i)$ that lowercases and trims $y_i$. Let $\mathrm{contains}(u,v)$ indicate that string $u$ contains string $v$ as a contiguous substring.
\[
h_{\text{sa}}(y_i; \mathbf{y}_i^*) =
\begin{cases}
0 & \text{if } \exists \text{ } y_i^* \in \mathbf{y}_i^* \ \text{s.t. } \mathrm{contains}(\mathrm{norm}_{sa}(y_i), \mathrm{norm}_{sa}(y_i^*)),\\
1 & \text{otherwise.}
\end{cases}
\]
If output includes additional text, grading is based on the presence of any acceptable answer string.

\section{Additional Experiments: Number of Candidate Responses vs. Black-Box UQ Performance}
\label{sec:num_responses}
\begin{figure}[H]
    \centering
    \includegraphics[width=\linewidth, height=0.8\textheight]{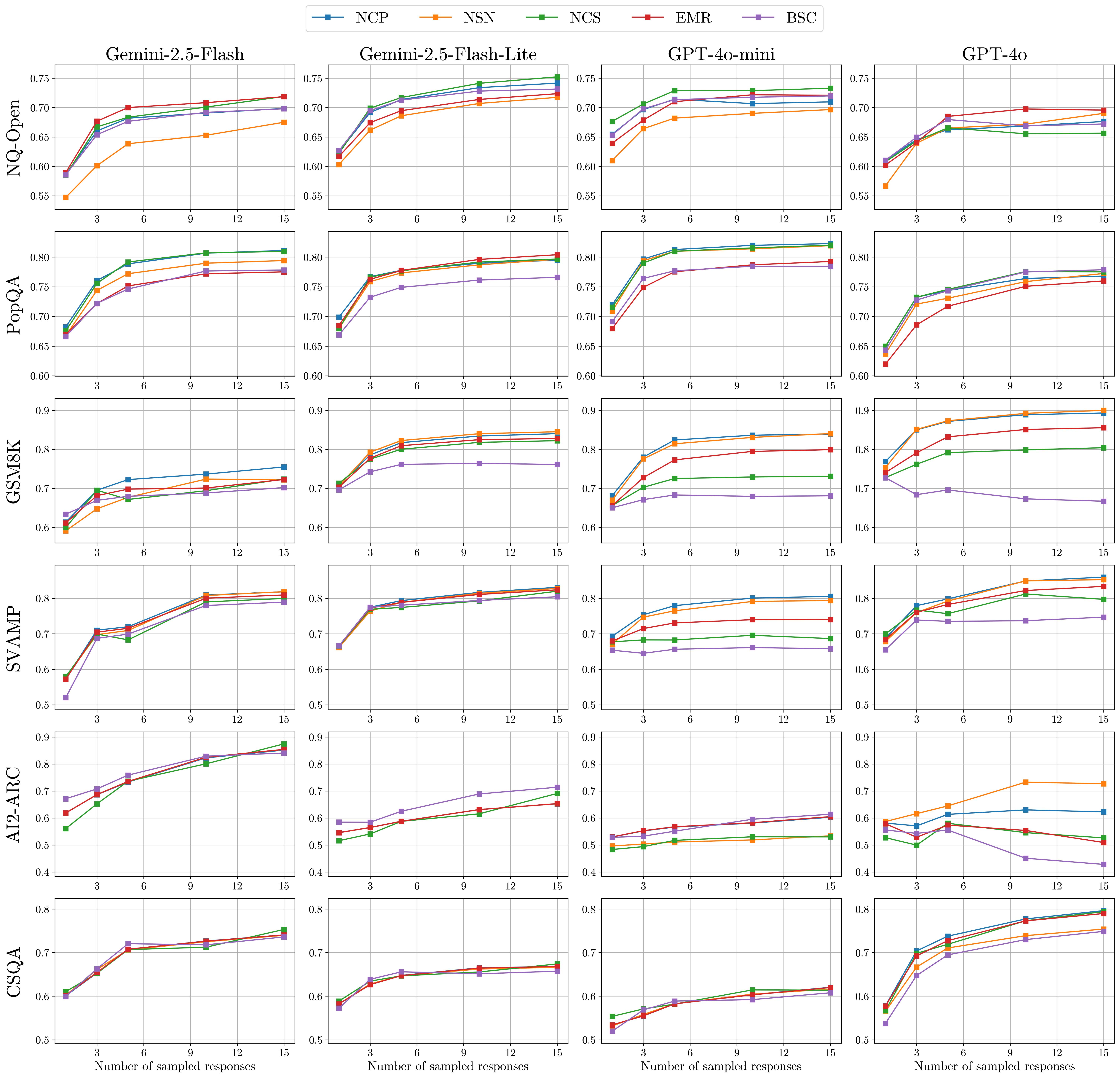}
    \caption{Hallucination Detection AUROC by Number of Sampled Responses}
    \label{fig:num_responses}
\end{figure}

To investigate the effect of number of candidate responses $m$ on the performance of black-box scorers, we re-compute all black-box confidence scores for $m=1, 3, 5, 10, 15$ for each of our 24 LLM-dataset scenarios. We compute the scorer-specific AUROC value for each value of $m$ and evaluate the impact of number of candidate responses on hallucination detection performance. These results are depicted in Figure \ref{fig:num_responses}.

Overall, the results indicate that hallucination detection performance of the black-box scorers improves considerably with number of candidate responses $m$. For instance, hallucination detection AUROC of the various black-box scorers on GPT-4o responses on CSQA improve from 0.54-0.57 with $m=1$ to 0.75-0.8 with $m=15$. In particular, these performance improvements occur approximately monotonically with diminishing returns to higher $m$, consistent with findings from previous studies \citep{kuhn2023semanticuncertaintylinguisticinvariances, manakul2023selfcheckgptzeroresourceblackboxhallucination, lin2024generatingconfidenceuncertaintyquantification, Farquhar2024}. This general trend is consistent across all 24 scenarios, with only a few exceptions, notably BSC for GPT-4o responses on GSM8k, and BSC, NCS, and EMR for GPT-4o responses on AI2-ARC.

\end{document}